\documentclass[letterpaper]{article}

\usepackage{natbib,alifeconf}  

\usepackage{subcaption} 
\usepackage{hyperref}   
\usepackage[dvipsnames]{xcolor}     
\usepackage{amsmath}                
\usepackage{siunitx}

\definecolor{dark-red}{rgb}{0.4,0.15,0.15}
\definecolor{dark-blue}{rgb}{0.15,0.15,0.8}
\definecolor{medium-blue}{rgb}{0,0,0.5}
\hypersetup{
    colorlinks, linkcolor={dark-red},
    citecolor={dark-blue}, urlcolor={medium-blue}
}

\usepackage{tikz}

\newcommand\copyrighttext{%
  \footnotesize  \textcopyright Accepted at the 2023 Conference on Artificial Life (ALife)}
\newcommand\copyrightnotice{%
\begin{tikzpicture} [remember picture,overlay]
 \node[anchor=south,yshift=10pt] at (current page.south)
    {\copyrighttext};
\end{tikzpicture}%
}

\definecolor{2check}{rgb}{0.4,0.0,0.1}
\definecolor{NOTE}{rgb}{1,0.0,0.1}

\title{ALIFE2022 template}

\title{Individuality in Swarm Robots with the Case Study of Kilobots:\\
Noise, Bug, or Feature?}
\author{Mohsen Raoufi$^{1,2,3}$, Pawel Romanczuk$^{1,2}$ \and Heiko Hamann$^{1,4}$
\\
\mbox{}
\\
$^1$Science of Intelligence, Research Cluster of Excellence, 10587 Berlin, Germany
\\
$^2$Institute for Theoretical Biology, Department of Biology, Humboldt Universität zu Berlin, Berlin, Germany
\\
$^3$Department of Electrical Engineering and Computer Science, Technical University of Berlin, Berlin, Germany
\\
$^4$Department of Computer and Information Science, University of Konstanz, Konstanz
\\
mohsenraoufi@icloud.com}

\begin{document}
\maketitle


\begin{abstract}
Inter-individual differences are studied in natural systems, such as fish, bees, and humans, as they contribute to the complexity of both individual and collective behaviors. However, individuality in artificial systems, such as robotic swarms, is undervalued or even overlooked. Agent-specific deviations from the norm in swarm robotics are usually understood as mere noise that can be minimized, for example, by calibration. We observe that robots have consistent deviations and argue that awareness and knowledge of these can be exploited to serve a task. We measure heterogeneity in robot swarms caused by individual differences in how robots act, sense, and oscillate. Our use case is Kilobots and we provide example behaviors where the performance of robots varies depending on individual differences. We show a non-intuitive example of phototaxis with Kilobots where the non-calibrated Kilobots show better performance than the calibrated supposedly ``ideal" one. We measure the inter-individual variations for heterogeneity in sensing and oscillation, too. We briefly discuss how these variations can enhance the complexity of collective behaviors. We suggest that by recognizing and exploring this new perspective on individuality, and hence diversity, in robotic swarms, we can gain a deeper understanding of these systems and potentially unlock new possibilities for their design and implementation of applications.
\end{abstract}

\section{Introduction}
\copyrightnotice
While in artificial swarms, such as swarm robotics, heterogeneity in software and hardware is only appreciated since recently, the concept is widely recognized in studies of natural (complex) systems, such as fish schools, animal groups, and humans.
We first give examples of individuality in natural systems and then describe how it is viewed in examples of artificial systems.
\newcommand\figTwoHeight{1.1}
\begin{figure*}[!t]
\centering
    \subcaptionbox{}{\includegraphics[width=\figTwoHeight in,angle=-90]{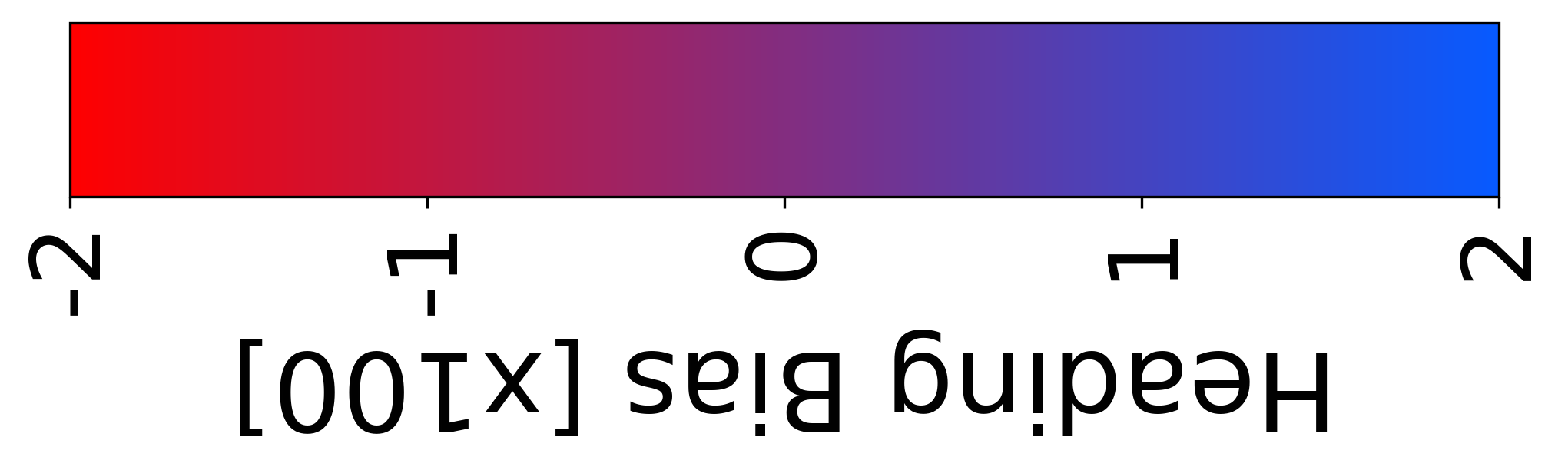}}
    \hfill
    \subcaptionbox{}{\includegraphics[height=\figTwoHeight in]{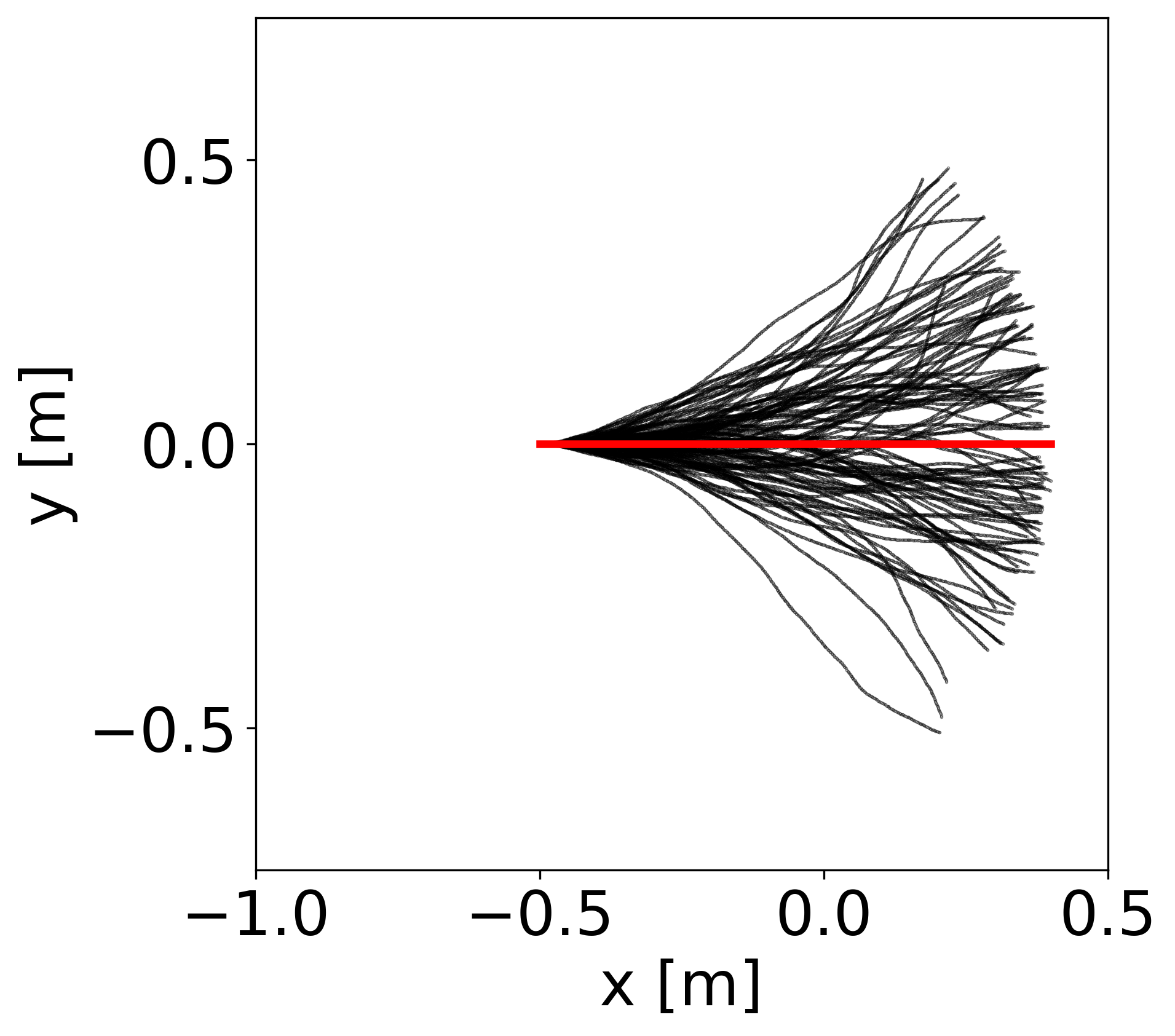}}%
    \hfill
    \subcaptionbox{}{\includegraphics[height=\figTwoHeight in]{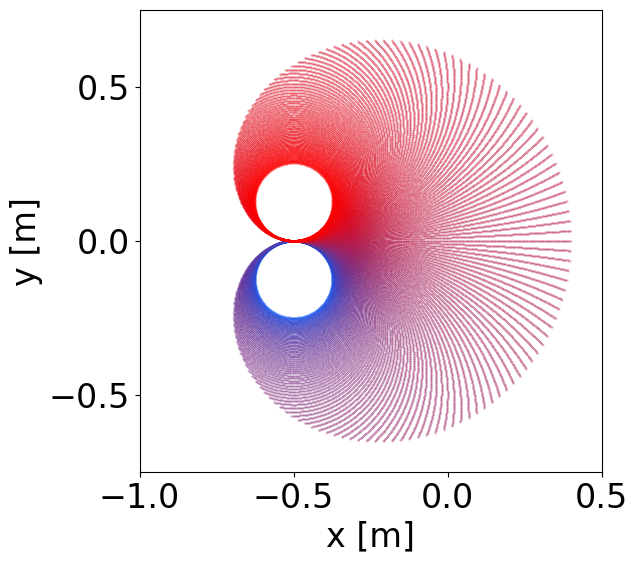}}%
    \hfill
    \subcaptionbox{}{\includegraphics[height=\figTwoHeight in]{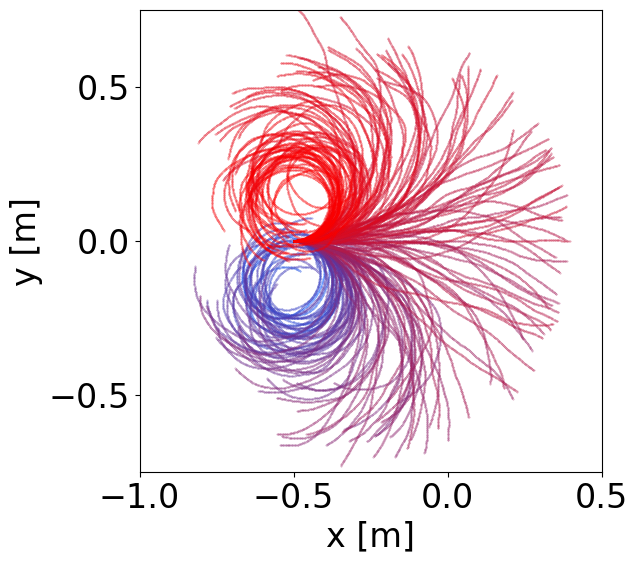}}%
    \hfill
    \subcaptionbox{}{\includegraphics[height=\figTwoHeight in,angle=0]{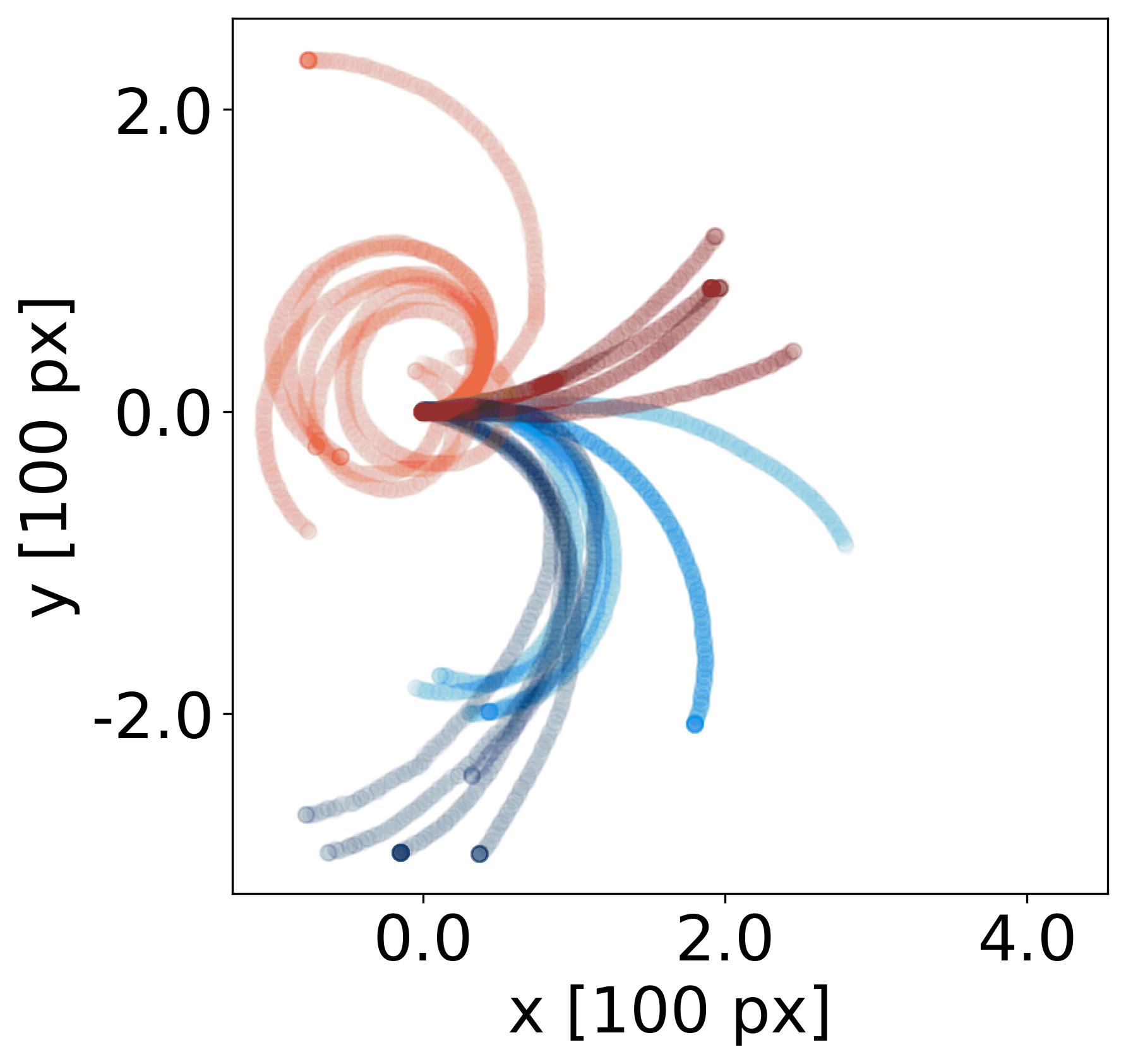}}%
    \hfill
    \subcaptionbox{}{\includegraphics[height=\figTwoHeight in]{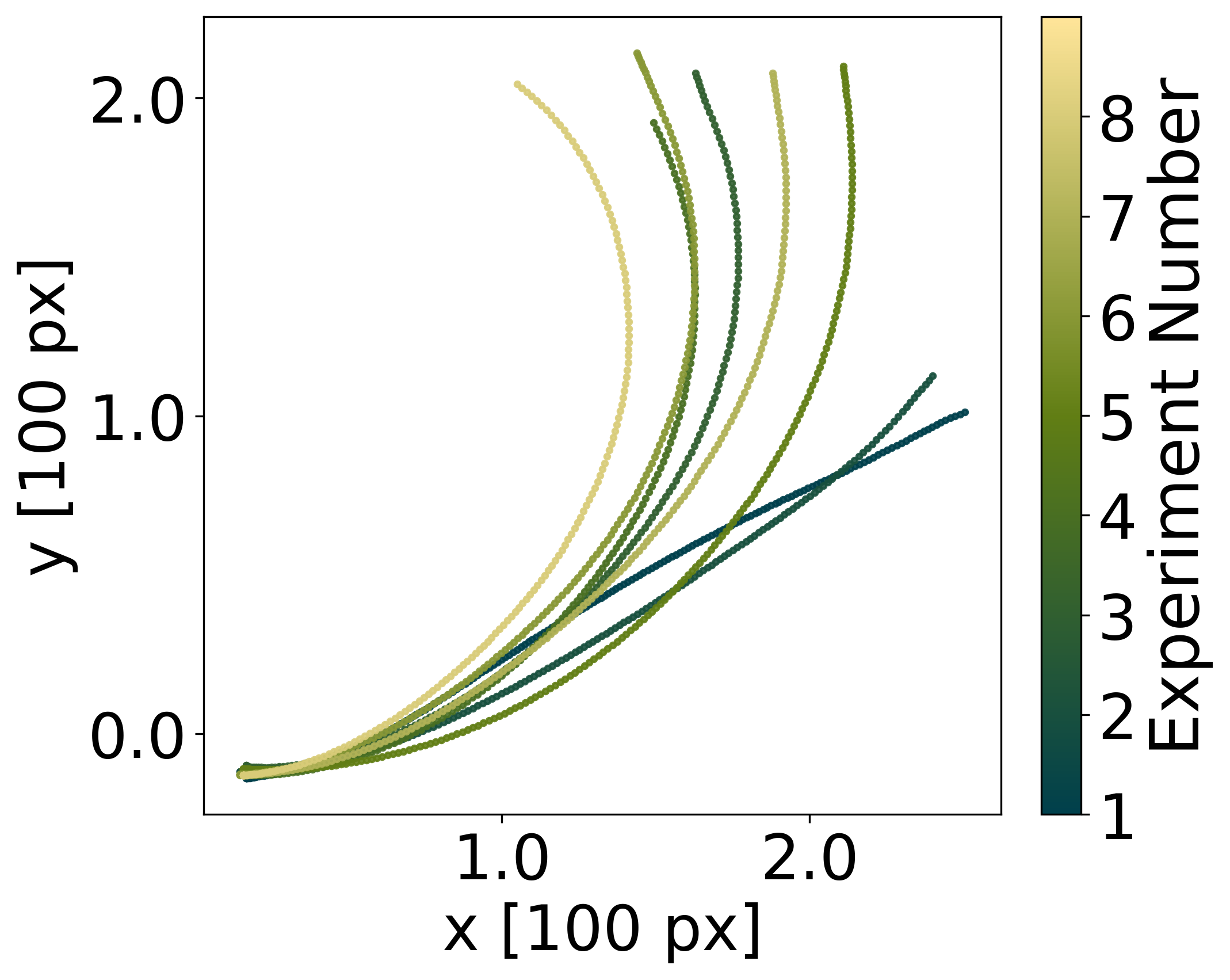}}%
\caption{Trajectories of robots moving on a straight line (b-d are in a simulation, e, f are real robot experiments). a) The color map of heading bias, showing the left- and right-biased robots in red and blue, respectively (used only in figures c, d). b) The red line is the trajectory of 100 ideal, identical robots without noise. The black lines are different realizations of the trajectories of identical robots with Gaussian noise. c) The noiseless trajectories of heterogeneous robots with a uniform heading bias distribution. d) The trajectories of heterogeneous robots with noise. e) Initial results of 4 Kilobots moving on a supposedly straight line for 4 independent repetitions starting at (0,0). Each color corresponds to a distinct robot. f) Trajectories of one robot (with ID number: 33784) moving forward for 8 different experiment trials starting at (0,0). 
}
\label{fig:forward_trajectories}
\end{figure*}
%
%


\subsection{Diversity and complexity in natural systems}
Diversity plays a significant role in the complexity of collective systems. The interplay of diversity and complexity in collectives is relevant in a variety of disciplines, such as physics, biology, economics, social science, and neuroscience, indicating a possible generality of the subject. According to~\citet{page2010diversity}, three different types of diversity are distinguished as: ``\textit{variation within} a type, differences \textit{across} types, differences \textit{between} communities." We focus on the first type of diversity, aka inter-individual variation, in this paper. 
Nature across many systems increases and maintains  diversity among individuals~\citep{ravary2007individual}. These variations are not only limited to physiological, or morphological differences but include behavioral and opinion diversities as well~\citep{del2018importance}.
\par
From a number of studies, we know that the complexity of system behaviors can stem from diversity~\citep{page2010diversity}. Fish are a well-studied species in this regard. For example,~\citet{nakajima2007righty} studied fish and the development of differences in their left-side body muscles compared to those of the right-side. They report that the ``righty" fish are more likely to be hooked on the right side of their mouth. Similarly, \citet{liu2009fish} studied the asymmetric development of motor functions in fish. \citet{laskowski2022emergence, ehlman2022developmental} also studied the development and emergence of individuality for twin female colonial fish. 
\citet{seeley2010honeybee} studied decision-making in bees, where the diversity of many individuals searching for a solution increases the chance of finding new options for nesting. Other studies focused on humans, rats, other social animals, and insects~\citep{rivalan2011inter, jeanson2014interindividual, lonsdorf2017more, jeanson2019within, ward2019individual, simmatis2020statistical}. 
The effect of variations on the performance of collective systems largely depends on the task. For example, in collective decision-making, some level of variation decreases the collective bias, but more variations do not necessarily improve its accuracy~\citep{raoufi2021speed}.
\subsection{Heterogeneity in Artificial Systems}
%
%
%
%
%
Different from natural systems, the inter-individual variations in artificial systems are often overlooked. As mentioned above, we study the first type of diversity~\citep{page2010diversity}. However, other types have recently received attention, for example, the diversity in the composition of the population, that is, having different robotic platforms (species) within a collective~\citep{prorok2017impact, dorigo2020reflections}. In this paper, we focus on inner-platform inter-individual differences, the so-called quasi-homogeneity~\citep{hamann2018swarm}. We even focus narrower, by excluding controllable or programmable variations (software heterogeneity); for example, robots with different control software, specialized in different tasks as by~\citet{dorigo2021swarm}. But rather we explore the intrinsic variations, that come naturally with the embodiment of robots and are an inseparable part of these systems.
\par
In most studies, the system behavior that emerges from the agent-agent, and agent-environment interactions is already complex, so that assuming a homogeneous system is sufficient~\citep{camazine2020self}.
\hyphenation{error}
The simplifying assumption of homogeneity in artificial systems, and in particular swarm robotics, improves tractability. We divide such assumptions into two main groups: noise and error. 
For the first group, individuality is seen as agents being deviated from the collective \textit{norm}. To deal with this matter in the modeling, one increases the variation of the noise to the extent that it covers the inter-individual variation, resulting in an increase of (aleatoric) uncertainty in the model~\citep{valdenegro2022deeper}, which is indeed due to the (epistemic) uncertainties of the system that is seemingly unknown to the observer. We highlight the possibility to extract information from this ``noise" that can be exploited and help us predict the behavior of the system more accurately. We use the example of heading bias for real Kilobots~\citep{rubenstein2012Kilobot} to elaborate more upon the concept. We argue that individual robots show persistent non-zero bias whose time correlation is infinitely large, which makes the noise assumption questionable.
Another engineering solution followed by the noise assumption is the attempt to calibrate robot sensors and motors. 
Although it reduces variations, the effect is only temporary and often deteriorates over time that is, calibrated robots eventually get decalibrated and deviate from the norm again. We ask: what is the acceptable extent to which an engineer should be concerned about the decalibration of robots? For the example of the heading bias of Kilobots in an optimization task, we show that the deviations from the ideal robot do not necessarily result in a performance decrease, but rather counter-intuitively enhance it in certain cases.
\par
The second approach is the regulation of deviations using control feedback \citep[e.g., ][]{wang2016safety}. By interpreting deviations from the \textit{desired} behavior as an ``error", which is meant to be regulated by the control system, the robot constantly tries to modulate its natural deviations and to minimize the error. This requires a feedback signal to form a closed loop~\citep{meindl2021collective}. However, in minimal swarm robots with simple noisy perceptions, and stochastic actuators the feedback solution is either expensive or impossible. We ask: should we treat individuality as noise, or a bug and hence try to solve it? Or is it rather a feature that the individual (or the collective) can exploit? Nature has shown a great ability to increase diversity, and to find a way to take advantage of it. Given that most of the swarm robotic systems are bio-inspired, it seems even mildly ironic to ignore or even \textit{fix} this feature.
\par
%
In the remainder of the paper, we introduce heading bias to the model of a moving agent in 2D space and accordingly modify the Kilobot simulator in ARGoS~\citep{pinciroli2012argos}. Then, we report our results on the heterogeneity of heading bias for Kilobots and how it develops over time. In the third section, we investigate the effect of heterogeneity in phototaxis and random walk. We compare the results of three different parameters for phototaxis and discuss how learning and evolutionary algorithms are influenced by individuality. Lastly, we measure, report, and discuss variations in sensing and internal frequency as other aspects of individuality. We also mention how collective decision-making and synchronization complexities stem from individuality, which will give an insight into future works, and together with the previous sections, open the stage for further investigation of the individuality in collectives.
\section{Heterogeneity in Motion}
In this section, we study how robots with heterogeneous motor abilities perform tasks differently. First, we model the motion of a differential-wheel robot and describe the effect of heading bias on its motion in simulation. Second, we report the data we analyzed from real Kilobot experiments, where robots are supposed to walk in a straight line.
\subsection{Model}
\begin{figure}[t]
\centering
    \includegraphics[width = \linewidth]{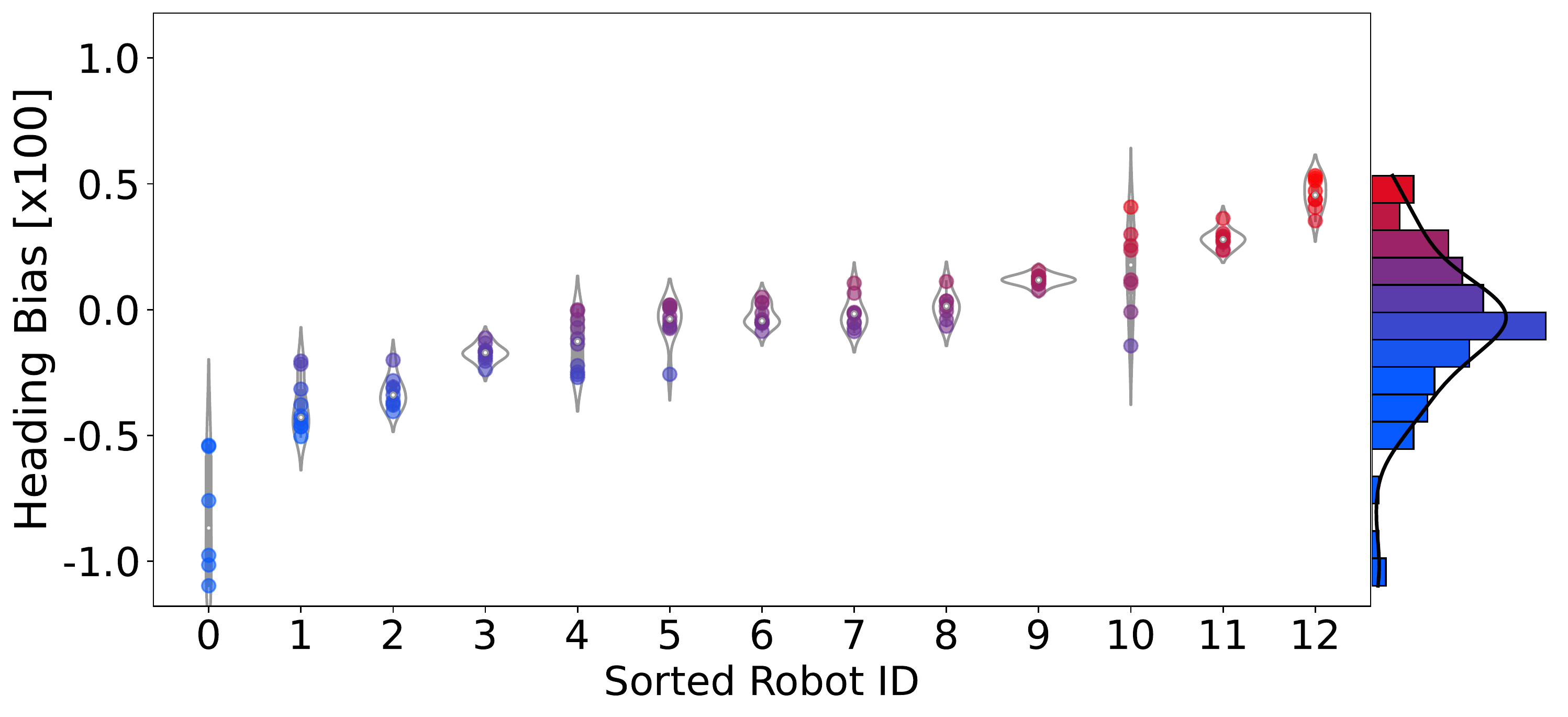}
\caption{Mean heading bias (over time) for each experiment grouped by robot ID, sorted by their mean heading bias, showing the individuality of heading bias for Kilobots. The right histogram is the ensemble distribution if we remove the dimension of individuality. 
}
\label{fig:head_bias_vs_id}
\end{figure}
%
%
%
%
%
Variations in actuation abilities among agents lead to different movement dynamics. We model a robot moving with speed of $|v|$ in a 2-dimensional $(x, y)$ space with a heading angle of $\theta$ using these equations of motion:
\begin{eqnarray}
    &\left[\begin{array}{c}\dot{x} \\ \dot{y}\end{array}\right]
    =
    \left[\begin{array}{c}{v}_x \\ {v}_y\end{array}\right]
    =
    |{v}| \left[\begin{array}{c}\cos(\theta) \\ \sin(\theta) \end{array}\right] , \nonumber
    \\
    &\dot{\theta} = \omega
    \label{eq:EoM_2D}
\end{eqnarray}
%
%
%
For a differential-wheeled robot, the component of the velocity perpendicular to the heading is zero. 
The remaining components are the linear and rotational velocities (${v}$ and $\omega$, respectively). Assuming noisy actuation, we map the right ($m_\text{R}$) and left motors ($m_\text{L}$) nominal angular rates to robot linear and rotational velocities 
using the corresponding coefficients ($c_v$ and $c_{\omega}$)
and add a noise term:
\begin{eqnarray}
{v} = c_v (m_\text{R} + m_\text{L}) + \eta_v, \nonumber \\
\omega = c_\omega (m_\text{R} - m_\text{L}) + \eta_\omega,
\label{eq:diff_wheel}
\end{eqnarray}
where $\eta_v$ and $\eta_\omega$ are the linear and rotational Gaussian noise, respectively. In an ideal case, the \textit{desired} and actual motor velocities are the same. However, in a real-world case it is well-known in robotics, the zero-error assumption could only be achieved by constantly regulating the error. In an open-loop system without any feedback, it is inevitable that the desired and the actual signals drift from each other.
Moreover, even for identical robots that are mass-produced, each motor of each robot might deviate from its nominal properties over time leading to actuation heterogeneity.
Furthermore, the differences in mass, inertia, unbalanced distribution of mass, or friction coefficient add to the variation in the dynamic of motion.
In this paper, we do not aim to explain the sources of these uncertainties, and the differences between the motors causing the problem, which might be cumbersome or difficult to measure. Instead, we focus on a higher abstraction at the level of the equation of motion (Eq.~\ref{eq:EoM_2D}), which is universal across systems, either artificial or natural.
Considering all the aforementioned factors causing the non-ideal motion of robots, we have the following equation of motion for each individual robot~$i$:
\begin{eqnarray}
    &\left[\begin{array}{c}\dot{x}^i \\ \dot{y}^i\end{array}\right]
    =
    (|{v}^i| + \eta_v) \left[\begin{array}{c}\cos(\theta^i) \\ \sin(\theta^i) \end{array}\right] , \nonumber
    \\
    &\dot{\theta^i} = \omega^i + \eta_\omega
    \label{eq:EoM_2D_noisy}
\end{eqnarray}
Studies showed that inter-individual variation in speed ($|{v}^i|$) among agents leads to complex collective motion~\citep{peruani2018cold, klamser2021impact}. 
For the case of Kilobots, \citet{pinciroli2018simulating} reported ``strong inter-individual variations" for linear speed and measured the variance of speed distribution (or equivalently $\eta_v$) for \textit{calibrated} Kilobots.
We focus on the rotational motion and heterogeneity in the heading bias of Kilobots.
\subsection{Heterogeneity in Heading Bias}
Here, we provide simple measurements of individuality in robots to prepare our more sophisticated study of its impact on behavior.
To measure the heterogeneity in heading bias, here we only consider the simple straightforward motion as an ideal motion, where the desired rotational velocity is zero ($\omega^i_\text{des}~=~0$). We conducted experiments with real Kilobots and in simulation using ARGoS simulator. We program robots to move in a straight line and log their position. For real robot experiments, we record videos and post-process the video frames using an object detection algorithm from OpenCV library~\citep{opencv_library}. For simulation, we use the Kilobot extension of ARGoS~\citep{pinciroli2018simulating} and modified it by adding the explicit heading bias to the code\footnote{\url{https://github.com/mohsen-raoufi/Kilobots-Individuality-ALife-23}}.
\par
If we choose to reduce the heading bias heterogeneity to mere noise, we assume the following stochastic differential equation for the turning rate of each individual~$i$:
\begin{eqnarray}
    &\dot{\theta^i} = \eta_\omega, \quad \quad \eta_\omega \sim \mathcal{N}(\mu,\,\sigma^{2}).
    \label{eq:dtheta_noisy}
\end{eqnarray}
Notice that statistical properties of $\eta_\omega (\mu,\,\sigma$) are without index~$i$ as they are meant as a population-wide one-fits-all model.
To show the trajectories from this type of model, we modified the simulator by adding Gaussian noise~$\mathcal{N}$ to the nominal speed of each motor ($m_\text{R}, m_\text{L}$, Eq.~\ref{eq:diff_wheel}). The result of such model is a correlated random walk (Fig.~\ref{fig:forward_trajectories}-b) that does not qualitatively cover all trajectories we observe in real robot experiments (Fig.~\ref{fig:forward_trajectories}-e). The assumption of a one-fits-all noise model results in a mismatch between model and reality. Fitting the data of real robots to this model, we get a joint (ensemble) distribution for heading bias with mean close to zero ($\mu \approx 0$) and relatively large variance (see far-right, rotated histogram in Fig.~\ref{fig:head_bias_vs_id}). This is similar to the distribution of speeds ($|v|$) reported in~\citep{pinciroli2018simulating}. With this model, we get a high variance (seemingly aleatoric uncertainty), that is indeed reducible only if we consider the individuality of robots as we do next.
Instead, if we allow each robot its individual (Gaussian) noise model, we get:
\begin{eqnarray}
    &\dot{\theta^i} = \eta^i_\omega, \quad \quad \eta_\omega^i \sim \mathcal{N}(\mu^i,\,(\sigma^i)^{2}).
    \label{eq:dtheta_noisy_ind}
\end{eqnarray}
The added dimension (raised index~$i$) to the parameter space enables us to model the individuality of each robot, which leads to lower aleatoric uncertainty. We show the distribution of heading bias for each robot in Fig.~\ref{fig:head_bias_vs_id}. Here, the data for robots with a tendency to turn to left is shown in red (blue for right). The data confirms our reasoning as robots have persistent, non-zero ($\mu_i \neq 0$) heading biases that are individual-specific. The variation of the heading bias across different experiments for each robot differs, that is, some robots are more consistent in their heading bias than others (differences across $\sigma^i$). Nonetheless, the consistency of heading bias for each robot suggests a strong inter-agent variation among Kilobots. Furthermore, these intra-individual variations are on average smaller than the ensemble variance ($\sigma^i<\sigma$). 
\par
To simulate the heading bias, we add a deterministic off-balancing term to the left and right motor speeds in Eq.~\ref{eq:diff_wheel}:
\begin{eqnarray}
\tilde{m}^i_\text{R} = m^i_\text{R} + \delta^i, \nonumber \\
\tilde{m}^i_\text{L} = m^i_\text{L} - \delta^i.
\label{eq:off_balanced_diff_wheel}
\end{eqnarray}
This non-zero rotational velocity generates circular trajectories. We illustrated the results of the simulations as a proof of concept in Fig.~\ref{fig:forward_trajectories}-d,~c with and without noise, respectively. We get trajectories of robots in simulation that have a similar curvature to their real robot experiments, which was not possible even by increasing the variance of the zero-mean noise in Eq.~\ref{eq:dtheta_noisy}. Each line shows the trajectory for a specific robot with a unique heading bias color-coded by the spectrum shown in Fig.~\ref{fig:forward_trajectories}-a. Extreme heading biases result in circles of small radii, that resemble non-calibrated Kilobots, similar to what we observe in real robot experiments of Fig.~\ref{fig:forward_trajectories}-e (the orange curves). 
\par
%
%
%
%
%
%
\subsection{Development of Individuality over Time}
The development of individuality in natural systems has already been linked to a variety of factors, such as environmental, social, and behavioral reasons. For artificial systems, the source of such developments in hardware individualities can be traced down to aging of mechanical components, such as fatigue; undergoing a major disturbance, such as damage; or simply the change in the energy source, to name but a handful of causes. For the heading bias of Kilobots in particular, we observe that over the time of experiments, robots that initially are well-calibrated lose their calibration and lose their ability to go straight. This de-calibration process is another reason why individuality emerges in synthetic systems and why calibration is not a lasting solution. Different platforms most likely have different timescales for losing their calibration. For Kilobots we observe the heading bias of a single robot changes slightly over the course of different repetitions. While for a more sophisticated robot, it may take longer. The results of an experiment with 8~repetitions are shown in Fig.~\ref{fig:forward_trajectories}-f. The robot moved on a rather straight line in the initial experiment (green line), whereas in the later experiments, the straight line started to bend toward the left side of the robot (yellow line). This is not meant to be a full study of the concept but to indicate the real-world effects.
\section{Example Scenarios with Motion}
\renewcommand\figTwoHeight{0.93}
\begin{figure}[b!]
\centering
    \subcaptionbox{}{\includegraphics[height=0.835 in]{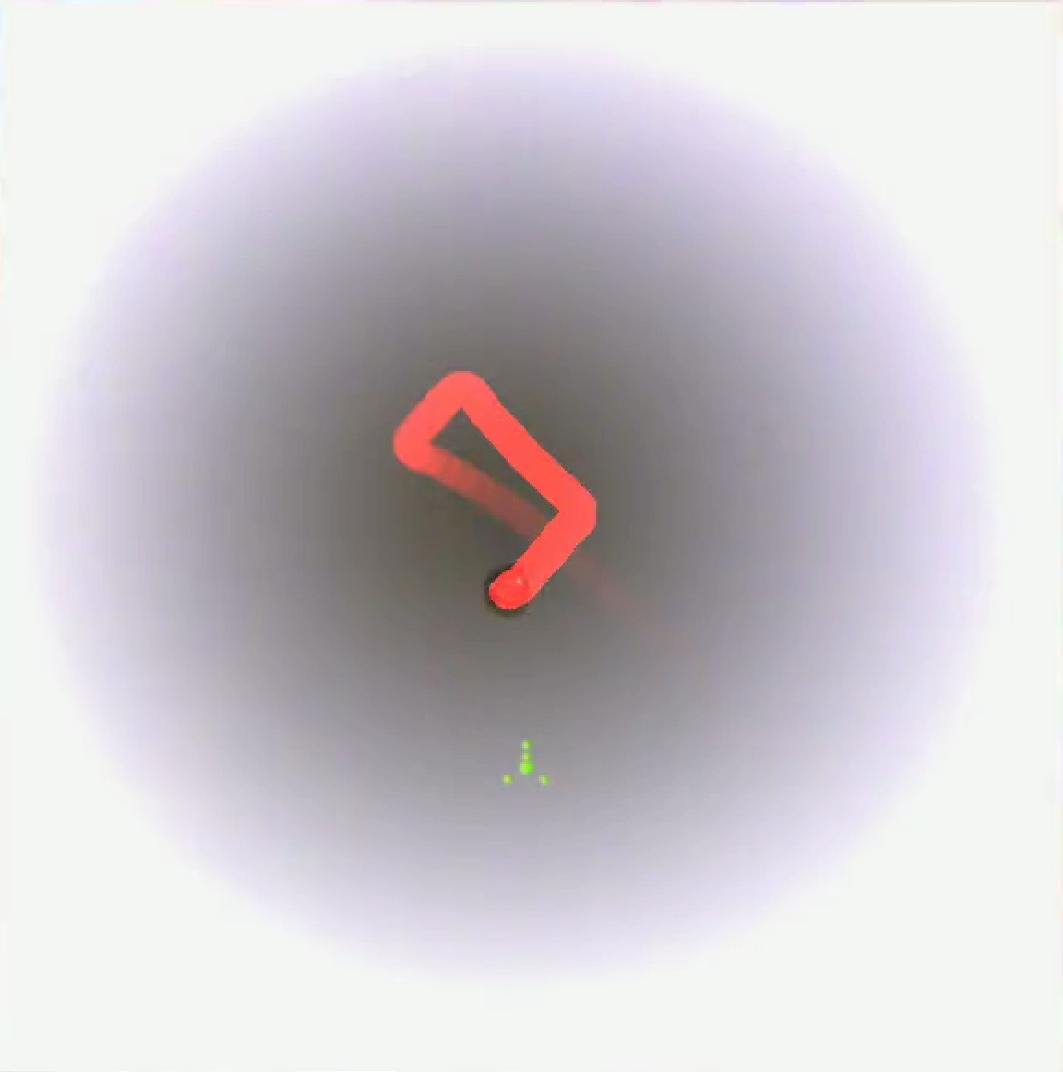}}
    \hfill
    \subcaptionbox{left-biased}{\includegraphics[height=\figTwoHeight in]{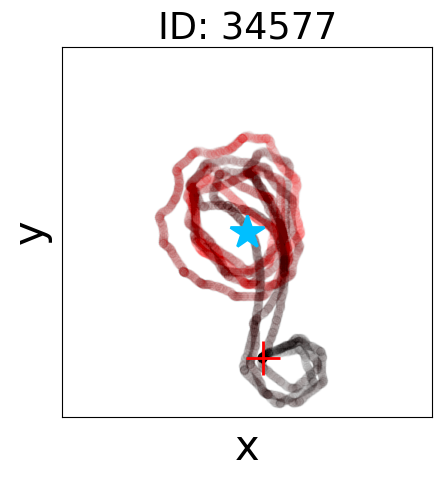}}%
    \hfill
    \subcaptionbox{ideal}{\includegraphics[height=\figTwoHeight in]{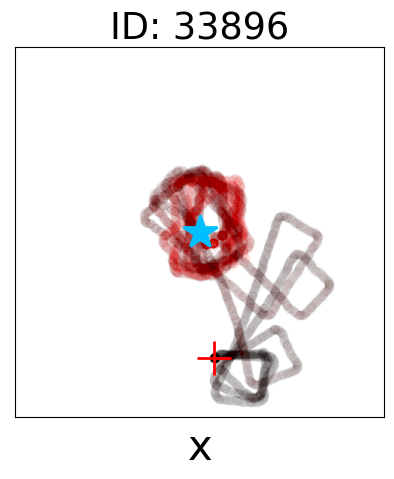}}%
    \hfill
    \subcaptionbox{right-biased}{\includegraphics[height=\figTwoHeight in]{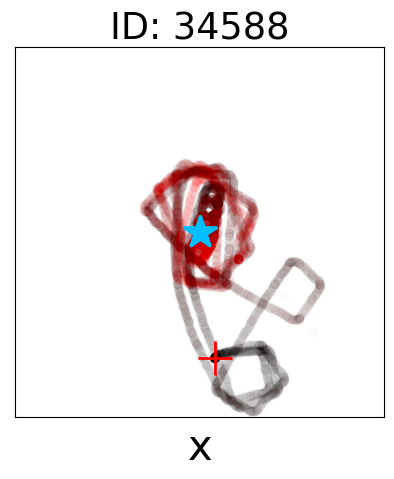}}%

\caption{Phototaxis with real Kilobots. a) A snapshot of one robot (same as in c) doing the deterministic phototaxis ($P_\text{R}=1.0$) around the center of the source, with the decaying red trace of its trajectory by post-processing the video. Without losing the generality of the results or algorithm, and for the sake of visualization, we inverted the light distribution so that the objective for the robot is to descend on light distribution. b-d) Trajectories of 3 robots with different heading biases, each showing 3 separate repetitions from the same initial point (red plus marker), doing phototaxis to reach the center of distribution (blue star marker). The points of trajectory are shown in more red over time. 
}
\label{fig:real_robot_phototaxis}
\end{figure}
To elaborate further on how individuality in motor abilities, and in particular heading bias, impacts the performance of robots, we pick two example behaviors, phototaxis (as exploitation), and random walk (as exploration). Phototaxis is a spatial sample-based optimization algorithm that maximizes the objective reward for the robot, which, in this case, is the light intensity distributed in a convex shape. The second example is designed for robots to do a random walk while gathering information from the environment. This example shows how different heading biases change the internal state of robots, namely the diversity or confidence of information gathered during the course of exploration.
%
%
\subsection{Phototaxis for Real Kilobots}
\label{sect:phottx_real_robot}
Phototaxis is an example behavior showing how simple organisms approach the center of an attractive light source~\citep{schmickl2010complex, baltieri2017active}. The algorithm is simple enough to be implemented on minimal robots, such as Kilobots. We added a light conductor on top of the light sensor of Kilobots as in our previous work~\citep{our_ICRA23} so that the individual robots can do point-wise sample-based phototaxis in space. 
Our real robot experiments prove that despite the simplicity of the algorithm and heterogeneity in motion, Kilobots can locate the center of the light source and exploit the reward.
%
%
\par
Our phototaxis algorithm is different from the random search explained in~\citep{pelkonen2018exploration} and the collective phototaxis as in~\citep{holland2019evolution}. Our more greedy algorithm boils down to the following procedure: if the intensity of the light sample gets closer to the objective intensity, the robot keeps going forward for a predetermined time duration, otherwise, it turns. The robot stops if it gets close enough (defined by a threshold) to the objective intensity. The turning direction is determined by a parameter~$P_\text{R}$ which is the probability to turn to the right (and $P_\text{L}=1-P_\text{R}$). To study how this parameter affects the performance of Kilobots, we consider three different configurations:

{\scriptsize$\bullet$} (asymmetric) deterministic turn to the right ({\small$P_\text{R}=1.0$}), 

{\scriptsize$\bullet$} symmetric stochastic turn to the left and right ({\small$P_\text{R}=0.5$}), 

{\scriptsize$\bullet$} and asymmetric stochastic turn to the left and right ({\small$P_\text{R}=0.25$}).
\\
Our experiments with real robots (see Fig.~\ref{fig:real_robot_phototaxis}) for the first algorithm suggest that robots have different performances (in approaching the source center). With $P_\text{R}=1.0$, a robot with a left heading bias (Fig.~\ref{fig:real_robot_phototaxis}-b) has a lower performance compared to the other robots that have either negligible (Fig.~\ref{fig:real_robot_phototaxis}-c) or right heading biases (Fig.~\ref{fig:real_robot_phototaxis}-d). In some cases, too strong left-biased robots failed to get closer to the center and left the area of interest.
%
%
%
%
\subsection{Phototaxis for Kilobots in Simulation}
\renewcommand\figTwoHeight{0.82}
\begin{figure}[b]
\centering
    \subcaptionbox{}{\includegraphics[height=\figTwoHeight in]{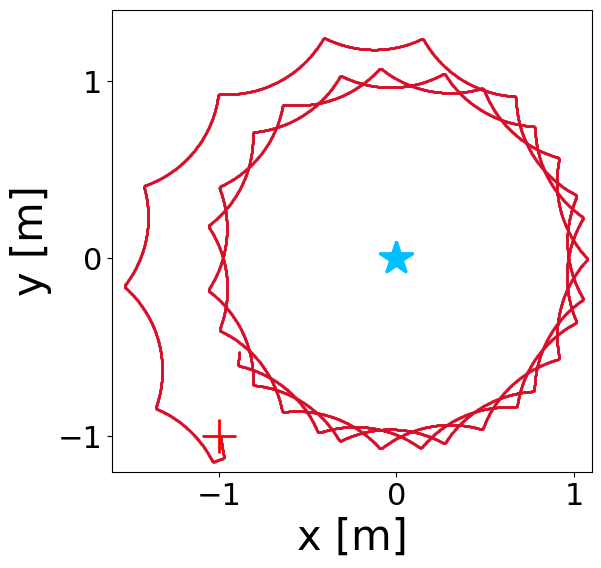}}%
    \hfill
    \subcaptionbox{}{\includegraphics[height=\figTwoHeight in]{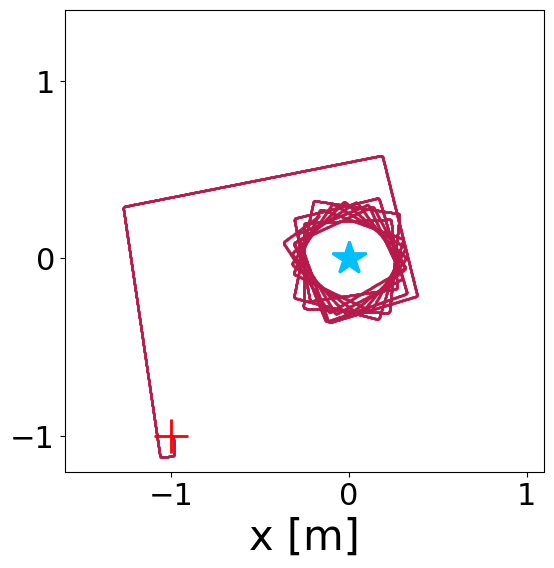}}%
    \hfill
    \subcaptionbox{}{\includegraphics[height=\figTwoHeight in]{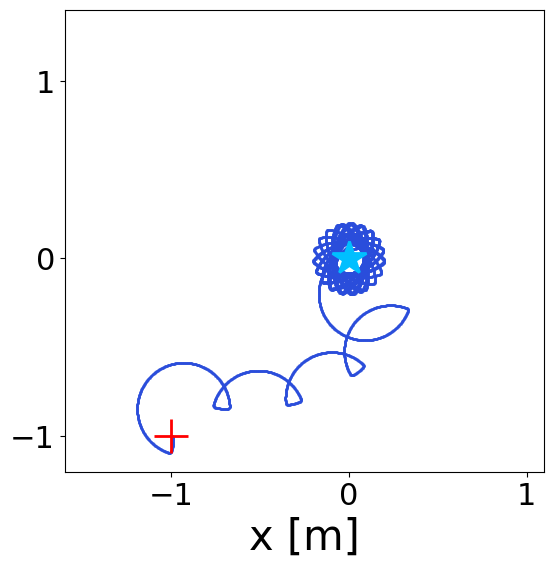}}%
    \hfill
    \subcaptionbox{}{\includegraphics[height=\figTwoHeight in]{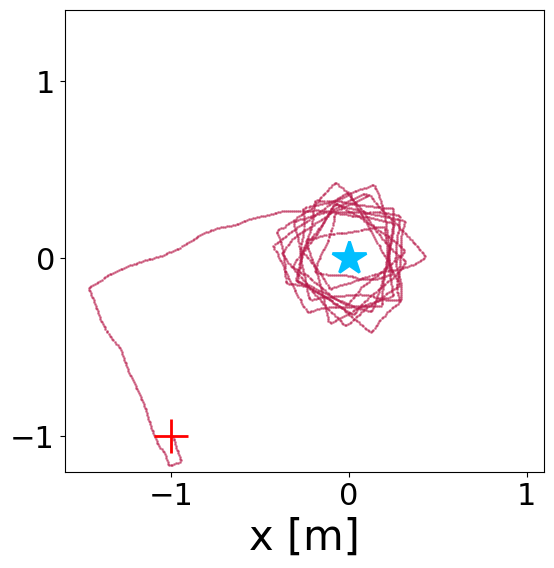}}%
    %
\caption{Simulated Kilobots doing phototaxis to the light source (blue star) at the center (0,0) for $P_\text{R}~=~1.0$. a-c) Trajectories of robots without noise for negative, zero, and positive heading-biased robots, respectively. d) The trajectory of a robot with noise and zero heading bias. 
}
\label{fig:phototaxis_traj_argos}
\end{figure}
\renewcommand\figTwoHeight{1.1}
\begin{figure*}[ht]
\centering
    \subcaptionbox{$P_\text{R} = 0.25$}{\includegraphics[height=\figTwoHeight in]{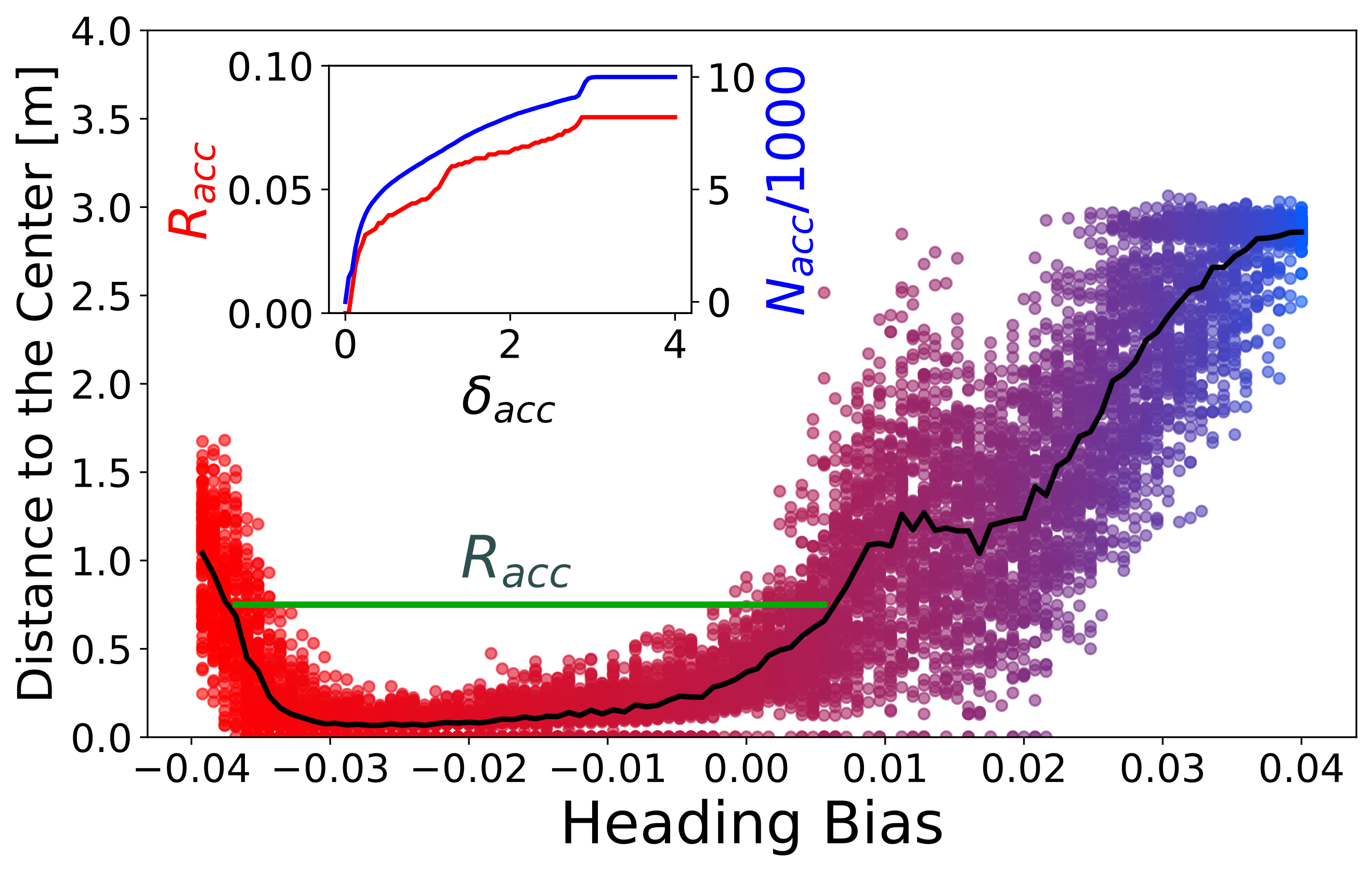}}
    \hfill
    \subcaptionbox{$P_\text{R} = 0.5$}{\includegraphics[height=\figTwoHeight in]{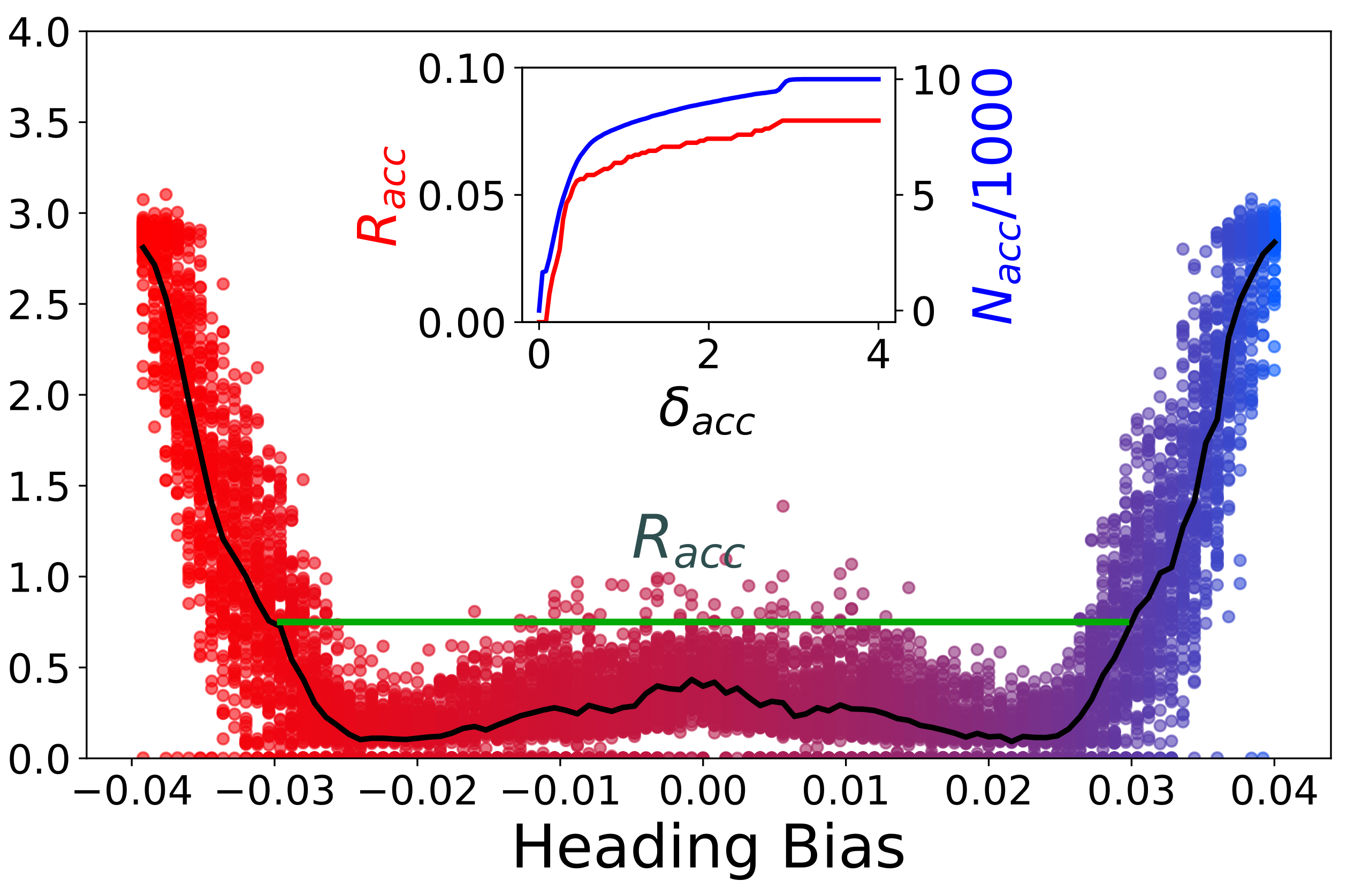}}%
    \hfill
    \subcaptionbox{$P_\text{R} = 1.0$}{\includegraphics[height=\figTwoHeight in]{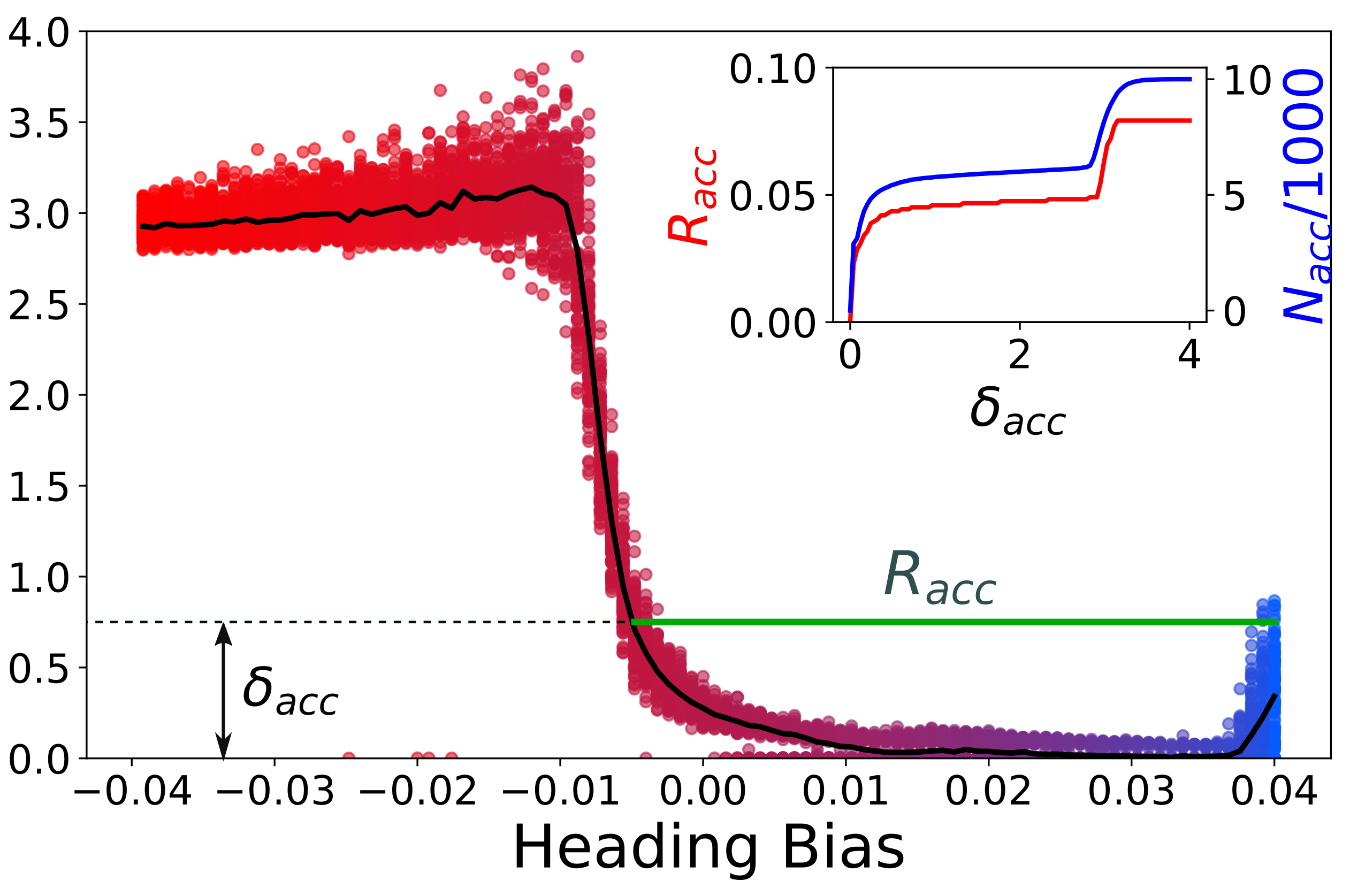}}%
    \hfill
    \subcaptionbox{}{\includegraphics[height=\figTwoHeight in]{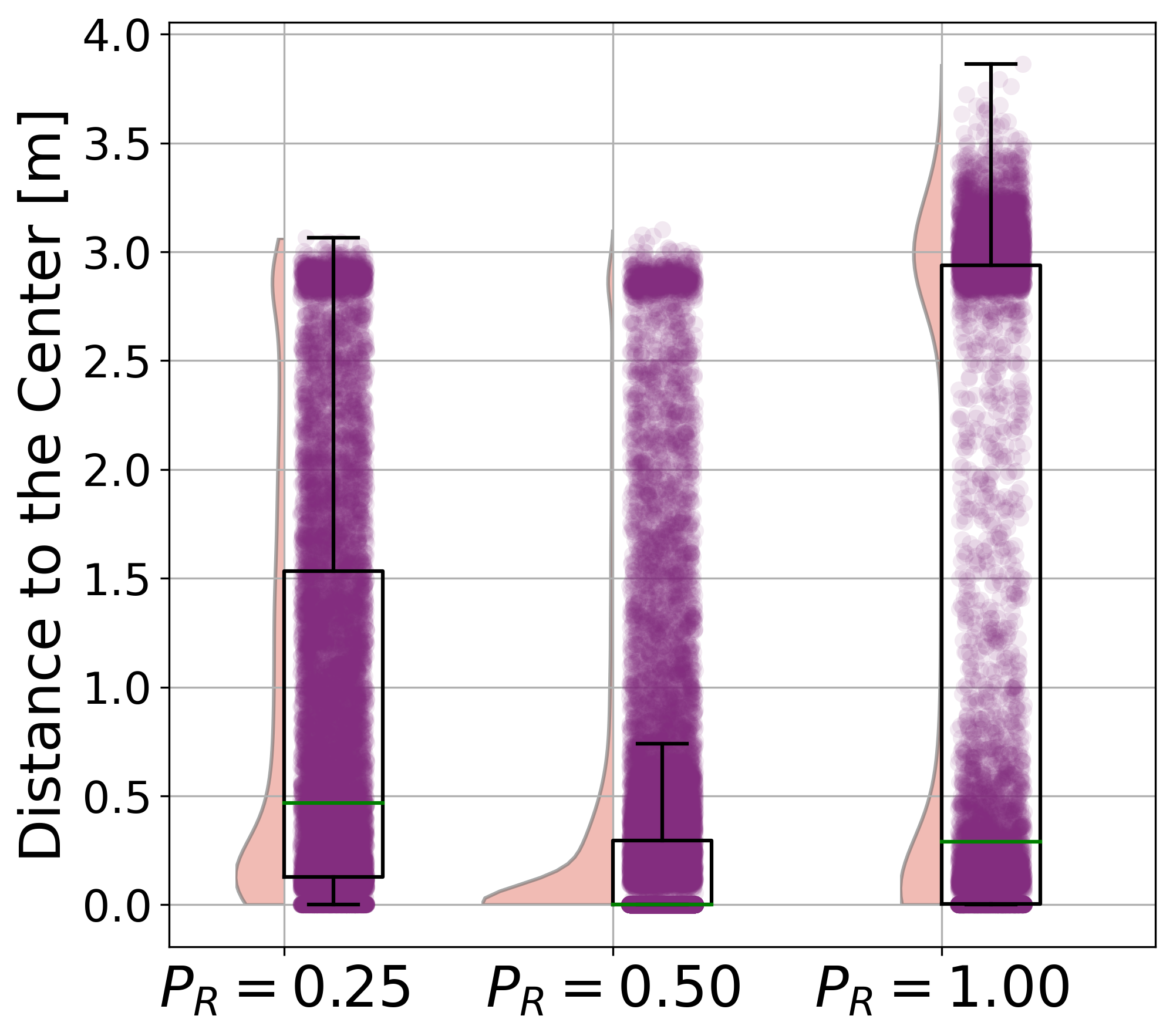}}%
\caption{Performance of robots with different heading biases in doing phototaxis with different parameters in simulation. a-c) A dot represents the performance of each robot at each simulation trial and is color-coded based on its heading bias. The black line shows the mean value over 100 Monte Carlo repetitions. The inset plots show $R_\text{acc}$ (red) and $N_\text{acc}$ (blue) vs thresholds $\delta_\text{acc}$. d) The ensemble distribution of all robots together (the light pink violin plot) by removing the individuality dimension. The black box plot shows the quantiles and the green line is the mean of all data points, each is shown by a purple dot.
}
\label{fig:phototaxis_dif_configs}
\end{figure*}
To study the effect of heading bias on the performance of phototaxis for Kilobots, we conduct experiments in ARGoS with the modified simulator we explained above. We test 100 simulated robots with heading biases uniformly distributed in the range of $[-0.04,0.04]$. Each robot executes the phototaxis algorithm, for 100 independent Monte Carlo simulations ($N_\text{MC}=100$). We distribute 100 (=$N_\text{MC}$) initial points and heading directions once for all robots, in order to make simulations comparable. Each experiment lasts for 200 seconds, with the time step of $0.1 \si{s}$.
We studied different simulation configurations with and without actuation noise (as in Fig.~\ref{fig:forward_trajectories}-c, d). We illustrated trajectories of 3 robots with different heading biases in Fig.~\ref{fig:phototaxis_traj_argos}. For the rest of the paper, we discuss the results of phototaxis with noise.
\par
We consider the distance of the robot to the center of the source for the performance metric as cost and calculate the average over the last 100 time steps. We illustrate the results for each trial as a point in Fig.~\ref{fig:phototaxis_dif_configs}. As expected, the robots vary greatly in their performance. This variation in the performance would have been otherwise ignored when assuming homogeneous robots.
A~key finding is that assumed ``perfect" robots without bias are, on average, outperformed by ``non-calibrated" robots (see Fig.~\ref{fig:phototaxis_dif_configs}-b, heading bias of $\pm 0.023$). 
This relates to our observation with real robots in Fig.~\ref{fig:real_robot_phototaxis}
, where the biased robot achieves more rewards.
To elaborate more on the optimality of non-zero bias robots, let us assume an evolutionary optimization algorithm that modifies the configuration of a robot (heading bias) over generations for a given fixed phototaxis parameter, e.g. $P_\text{R}=0.5$. The fact that the stable optimal heading biases are located at non-zero would cause the evolutionary algorithm to incline toward more biased configurations and select them more often over generations.
The attraction points depend on where to start the evolutionary optimization. For each of $P_\text{R}=0.25, 0.5$ there are two separate local optimums, one with a positive and the other with a negative heading bias.
This might give some insights into the development of diversity and heterogeneity in complex systems.
It confirms that calling individuality a \textit{bug} or a flaw (with negative impacts) is not always true. There are also other scenarios, where being biased causes harm in an asymmetric manner, for example, having an asymmetric chance of being hooked on one side of the body compared to the other for a righty fish~\citep{nakajima2007righty}.
%
%
\par
In addition, we compare the performance of different algorithms.
Each algorithm favors a specific range of heading biases.
For the deterministic algorithm (Fig.~\ref{fig:phototaxis_dif_configs}-c), where robots always turn to the right ($P_\text{R}=1.0$), the algorithm favors the right-biased robots more than the others. In comparison, for the algorithm with a higher chance to turn to the left ($P_\text{R}=0.25$, Fig.~\ref{fig:phototaxis_dif_configs}-a) the left-biased robots achieve higher rewards (lower cost). To highlight the effect of heterogeneity in optimization tasks we imagine a learning problem, where robots are supposed to learn the optimal value of~$P_\text{R}$. Given the results we provided here, it is predictable that robots with non-identical heading biases converge to different optimal parameters. A~left-biased robot would pick a lower $P_\text{R}$ compared to a right-biased robot. In that light, we argue that tuning one parameter for all robots by only optimizing the performance of a single robot with a specific feature (e.g., a non-biased robot) might not be the best practice.
\par
Another important point is the extent of acceptable ``uncalibratedness"; that is the range of heading bias within which robots perform reasonably well ($R_\text{acc}$). To quantify $R_\text{acc}$ we define a threshold ($\delta_\text{acc}$) for the performance, below which the criterion is satisfied. We show the acceptable range for $\delta_\text{acc}=0.75\ \si{m}$ in Fig.~\ref{fig:phototaxis_dif_configs}-a-c with the green horizontal line. Another related metric is $N_\text{acc}$, which counts the number of experiments (dots) performed below the threshold. 
We illustrated the performance metrics versus thresholds in the inset plots. We also compared the three algorithms in terms of the acceptable range (and number) versus the threshold in Fig.~\ref{fig:r_acc_vs_thresh}-a. With the highest threshold, all the dots have acceptable performance ($N_\text{acc}=10 \text{k}$). The ranking of the most efficient algorithms changes by decreasing the threshold. Specifically, we focus on the range [0, 1~m] (Fig.~\ref{fig:r_acc_vs_thresh}-b). 
The most efficient algorithm as of this metric depends significantly on where we set the threshold. Apart from providing performance comparison among algorithms, the acceptable range proposes some extent of freedom from calibrations. From an engineering point of view, not having to calibrate swarm robots would reduce the required effort, energy, and cost to maintain such large-scale systems.
\renewcommand\figTwoHeight{0.92}
\begin{figure}[b!]
\centering
    \subcaptionbox{}{\includegraphics[height=\figTwoHeight in]{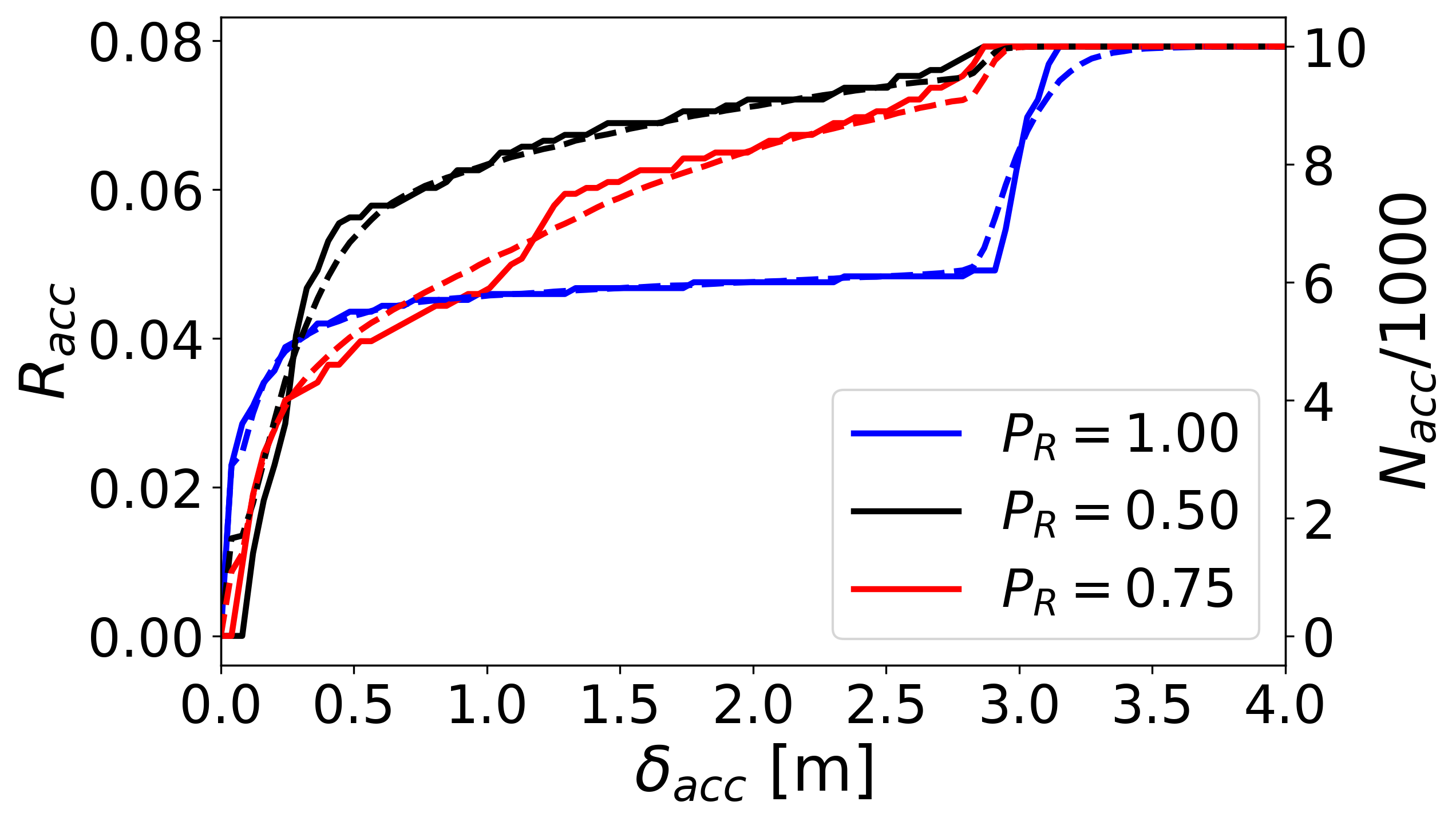}}%
    \hfill
    \subcaptionbox{}{\includegraphics[height=\figTwoHeight in]{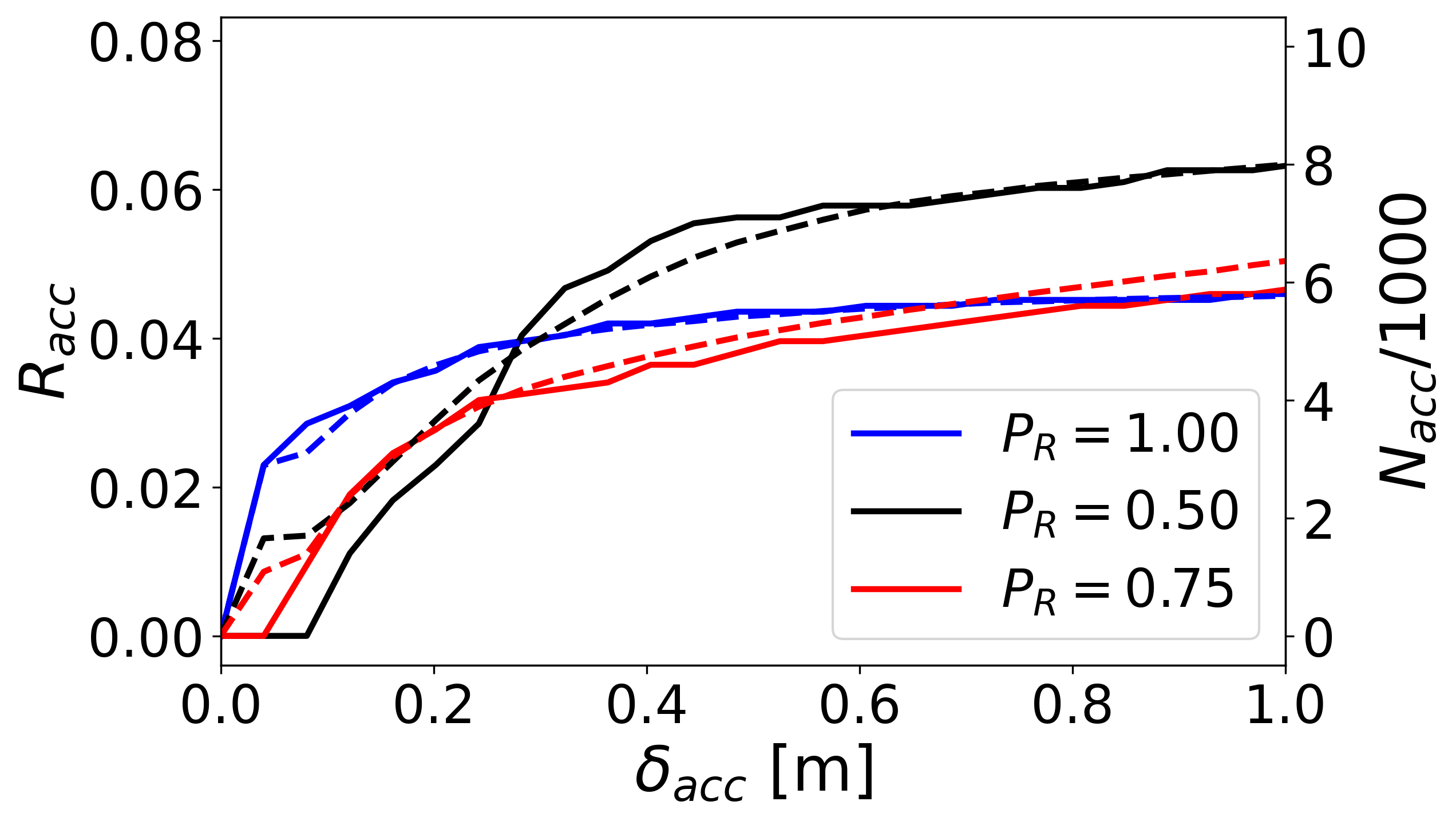}}%
\caption{Acceptable range $R_\text{acc}$ (solid lines), and the number of acceptable experiments $N_\text{acc}$ (dashed lines) for the three algorithms versus acceptable threshold $\delta_\text{acc}$. b) The same plot with limited range for $\delta_\text{acc}$ in [0,1].}
\label{fig:r_acc_vs_thresh}
\end{figure}
\par
If we ignore individuality, we get a joint distribution of performance for all robots (see Fig.~\ref{fig:phototaxis_dif_configs}-d.) Following this simplified interpretation we may draw conclusions that are either inaccurate or not generally valid. For example, the mean performance (denoted by the green line) represents a higher performance for $P_\text{R}=0.5$. However, this may not hold true for all robots. Also, the heavy upper tail for $P_\text{R}=1.0$ cannot be explained unless we look through the second dimension, which is individuality. This figure is an example showing the information that is neglected when ignoring the individuality dimension. 
\subsection{Random Walk}
Another example of a scenario for single robots with heterogeneous heading biases is the exploration task, during which they measure samples from the environment and gather information. Exploration plays an important role in decision-making, whether in individual or collective scenarios. Furthermore, the exploration correlates with the internal states of the robots, for example, their estimate of the environmental observable~\citep{ebert2020bayes, pfister2022collective, our_ICRA23}. In this section, we use the modified ARGoS simulator with the ability to capture the heterogeneity of heading biases for Kilobots in a bounded environment (see Fig.~\ref{fig:exploration}-a). The task for each robot is to perform a random walk and cover as much area as possible during its movement. To keep the generality of the performance metric, we only consider the non-overlapping area coverage regardless of the environmental distribution (Fig.~\ref{fig:exploration}-b). The area coverage measures the richness of information that a robot gathers during the exploration.
%
%
We conducted 100 independent Monte Carlo simulations for each of the 200 robots with different heading biases and illustrated the performance of robots in Fig.~\ref{fig:exploration}-c. The results suggest that the variation in robots' performance is significant. In this case, the ideal robot achieved the highest performance, while the biased ones gathered less diverse information. The richer information caused by higher area coverage leads to higher confidence. Consequently, robots with different heading biases will have heterogeneous certainty about the information they gathered during the exploration.
\par
In order to highlight the importance of heterogeneity in confidence, we consider the case of collective estimation where robots gather information from the environment and aggregate it to achieve a consensus, for example, on the mean value of the light distribution as by~\citet{ebert2020bayes} and~\citet{our_ICRA23}. We divide the aggregation methods into two main categories whether they take the confidence of information into the equation or not. The latter is usually referred to as na\"ive methods, for example, the DeGroot model for social learning~\citep{golub2010naive}. A~big family of example methods of the former, is the inference methods, such as Bayesian approaches~\citep{hkazla2021bayesian, chin2022minimalistic, pfister2022collective}. Despite the simplifying homogeneity assumption in na\"ive methods, collective scenarios using these methods manifest complex dynamics. Nonetheless, the heterogeneity of confidence among agents opens a new dimension to the complexity of information processing in collectives.
\renewcommand\figTwoHeight{1.0}
\begin{figure}[t]
\centering
    \raisebox{-0.5\height}{\subcaptionbox{}{\includegraphics[height=0.75 in]{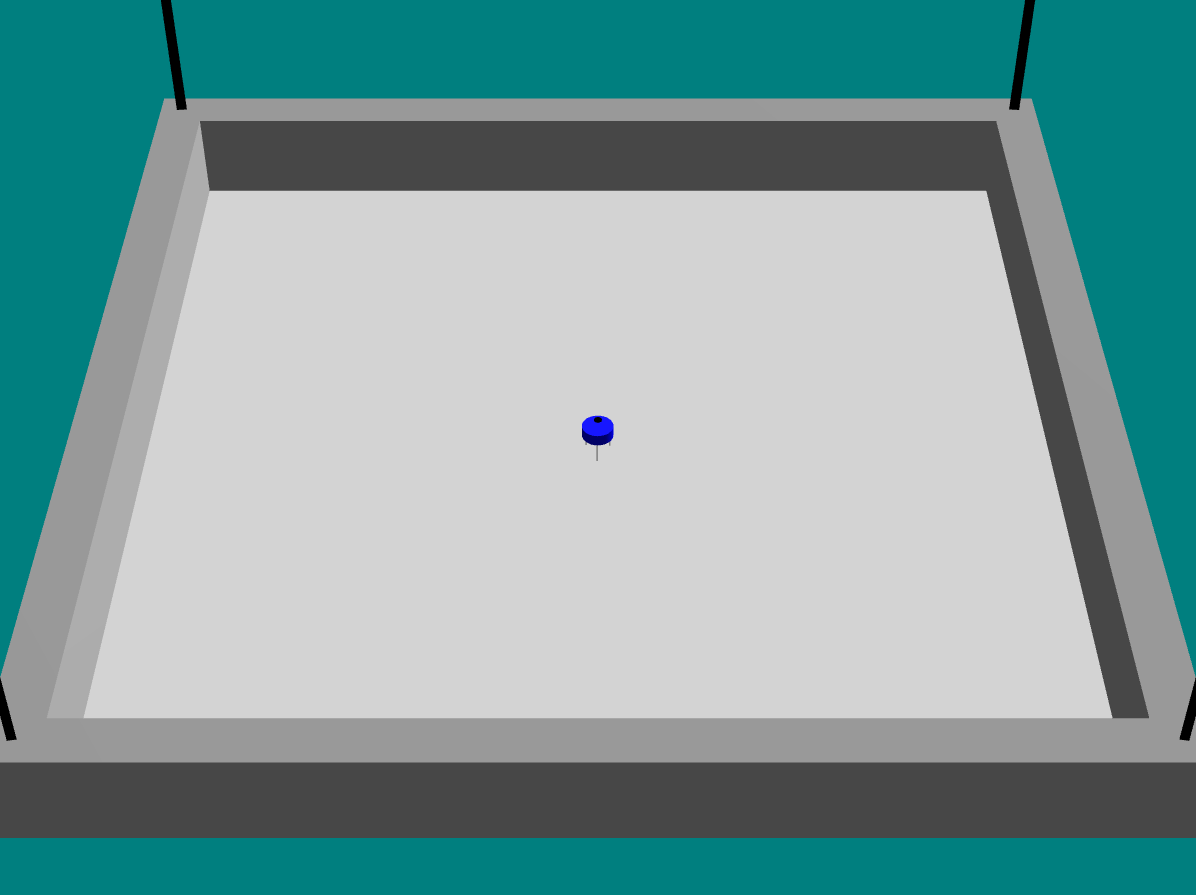}}}
    \raisebox{-0.5\height}{\subcaptionbox{}{\includegraphics[height=\figTwoHeight in]{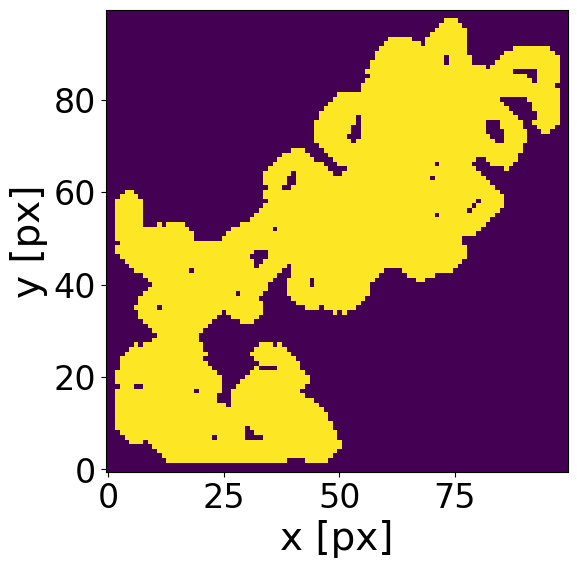}}}%
    \raisebox{-0.5\height}{\subcaptionbox{}{\includegraphics[height=\figTwoHeight in]{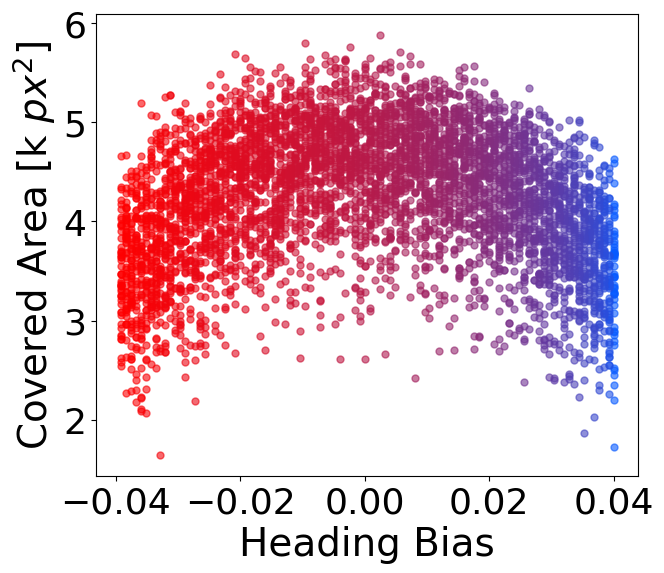}}}
\caption{Random walk for a single robot in a bounded environment. a) A snapshot of a Kilobot in ARGoS, b) an example of the area explored by a robot during exploration (yellow area), c) area coverage versus heading bias.}
\label{fig:exploration}
\end{figure}
\section{Other Aspects of Heterogeneity}
Besides heterogeneity in motion, we observe variation in other aspects of individual behaviors. In this section, we briefly describe, measure, and report the heterogeneity in sensing and natural frequencies of Kilobots.
\subsection{Heterogeneity in Sensing} 
Kilobots are equipped with a sensor that measures the ambient light intensity. Converted to a digital signal, the sensor output is a scalar variable in the range [0, 1023]. Although the precision of the sensor is considerably high, we observed and measured inter-individual variation among robots. To quantify the variation we designed an experiment as follows. We locate different robots exactly below the lens of a projector in a specific position and orientation. The projector is connected to a computer that controls the light intensity of a gray screen projected on the robot sensor. We change the light intensity of the screen by altering the so-called V-value of the gray color HSV. We sweep the V-value from zero to 1 (black to white), for four repetitions followed immediately by another, making a saw-teeth pattern (Fig.~\ref{fig:perception}-a). At the same time, the robot measures the intensity and sends it to the computer via serial communication.
\par
Using the periodicity of the signals, we trim the data of different experiments within a single period (see the red box in Fig.~\ref{fig:perception}-a). We compare the corresponding patch of different robots (Fig.~\ref{fig:perception}-b). The results confirm that inter-individual variation in light sensitivity among Kilobots is persistent and not negligible. For a Kilobot to perceive the surrounding world, it relies on its sensing through the single ambient light sensor. The difference in perceiving the state of the environment for agents can result in different decisions. For example, in a simple threshold-based binary decision, the inter-individual variation will cause disagreements among agents, even if the threshold is the same for all robots (Fig.~\ref{fig:perception}-c.)
%
%
%
%
\renewcommand\figTwoHeight{0.85}
\begin{figure}[t]
\centering
    \raisebox{-0.5\height}{\subcaptionbox{}{\includegraphics[height=\figTwoHeight in]{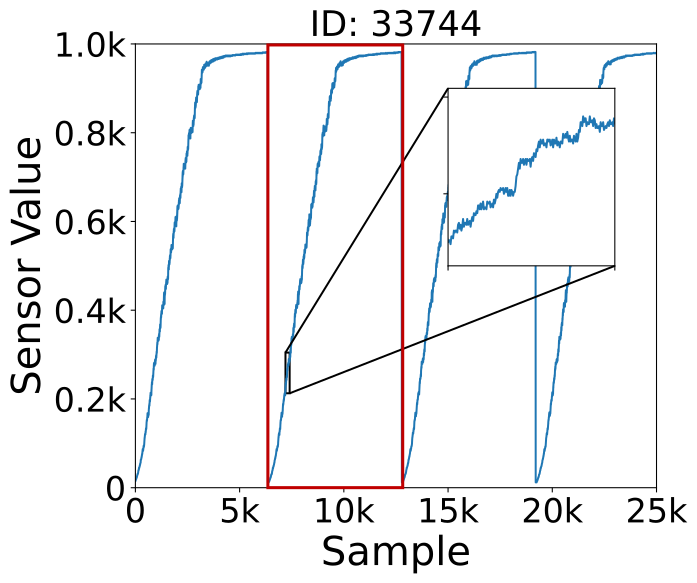}}}
    \hfill
    \raisebox{-0.5\height}{\subcaptionbox{}{\includegraphics[height=\figTwoHeight in]{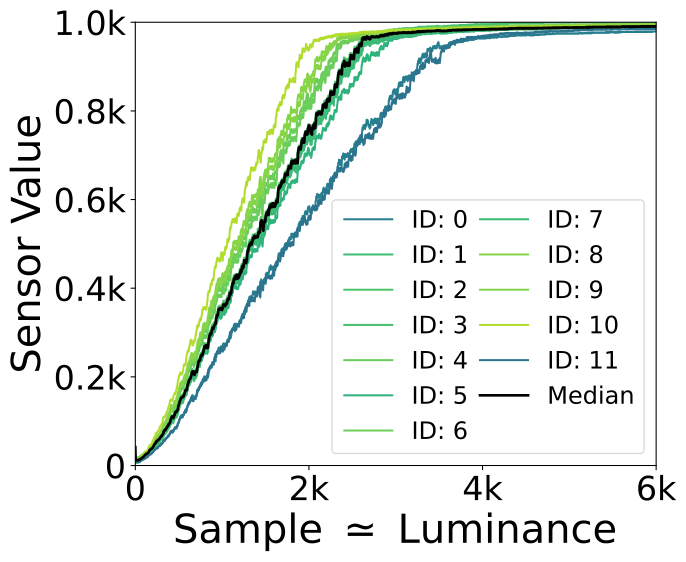}}}%
    \hfill
    \raisebox{-0.5\height}{\subcaptionbox{}{\includegraphics[height=\figTwoHeight in]{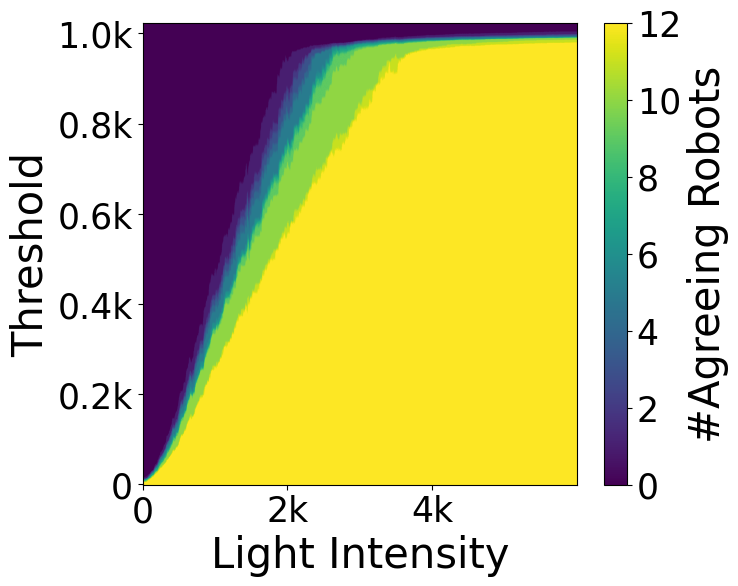}}}%
    %
\caption{Kilobot measurement of light intensity. a) The sensor value versus sample. b) 12 robots with  heterogeneous light sensitivity reading the same light intensity. The black line shows the median over different robots. c) The number of agreeing robots in a specific light intensity for different decision-making thresholds.
}
\label{fig:perception}
\end{figure}
%
%
%
%
%
\subsection{Heterogeneity in Natural Frequency}
\renewcommand\figTwoHeight{0.75}
\begin{figure*}[t!]
\centering
    \raisebox{-0.5\height}{\subcaptionbox{}{\includegraphics[height=0.7 in]{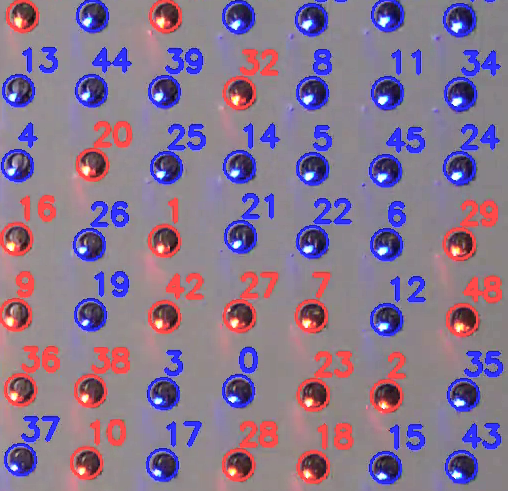}}}%
    \hfill
    \raisebox{-0.5\height}{\subcaptionbox{}{\includegraphics[height=0.75 in]{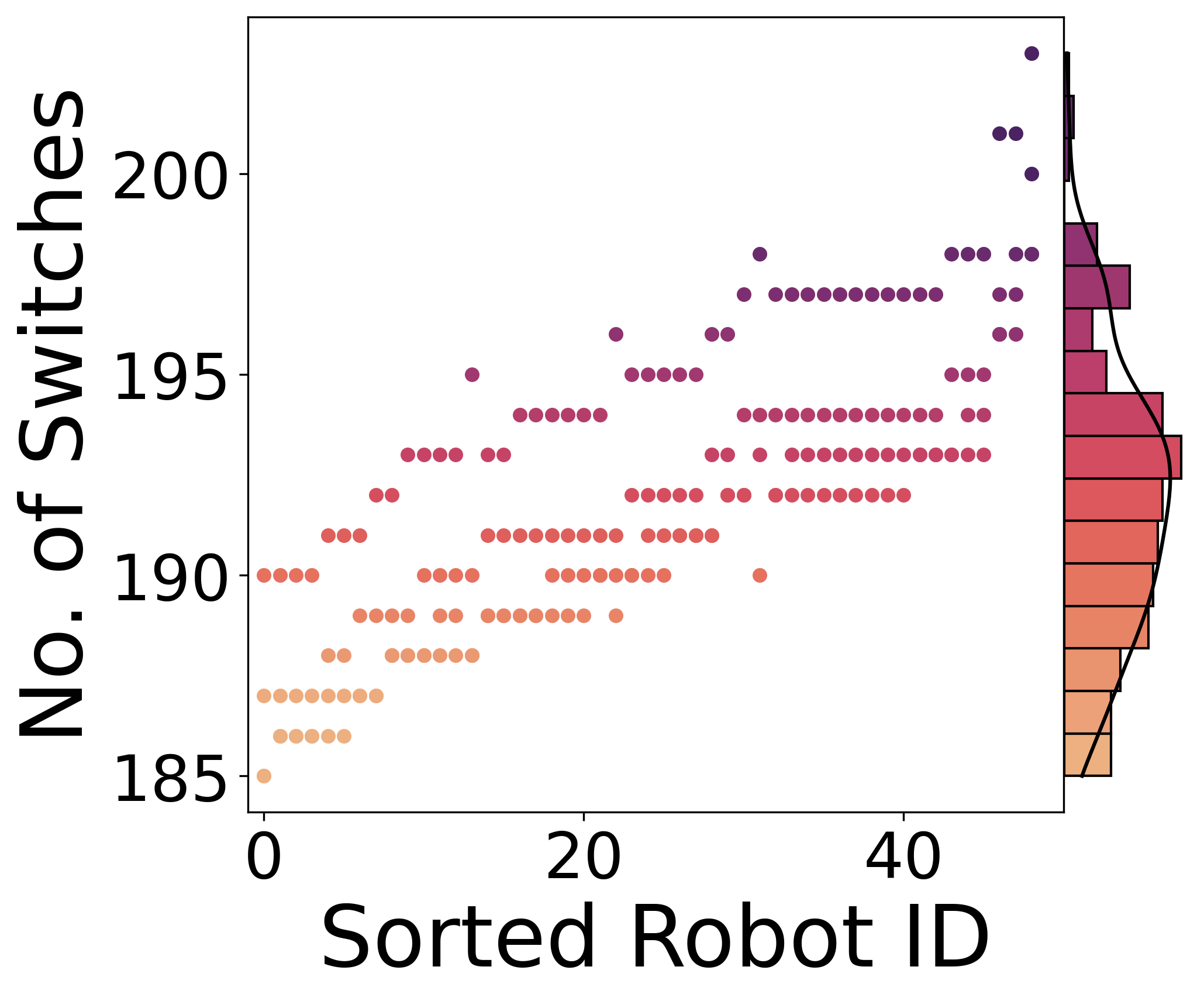}}}
    \hfill
    \raisebox{-0.5\height}{\subcaptionbox{no correlation}{\includegraphics[height=0.75 in]{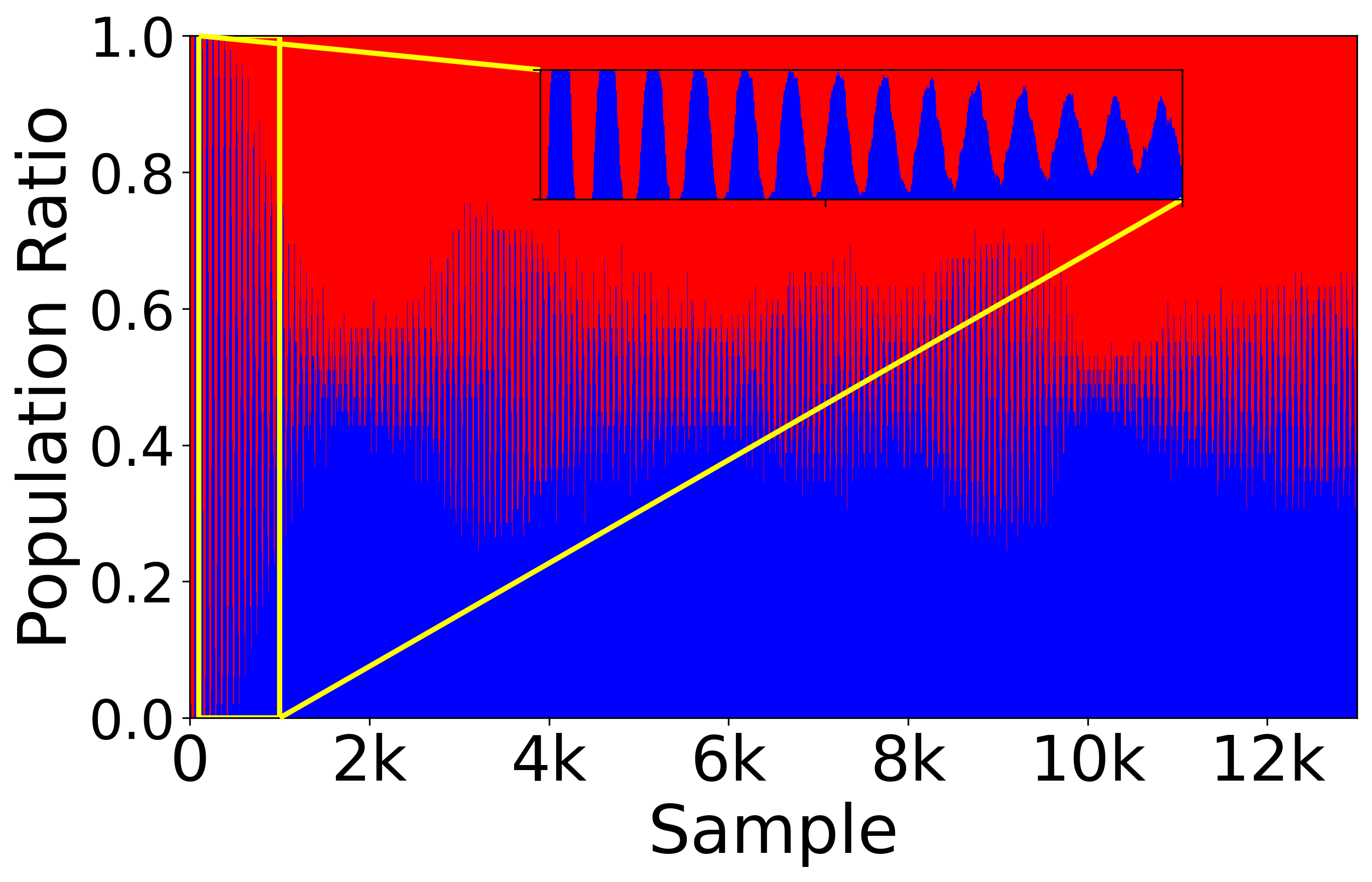}}}
    \hfill
    \raisebox{-0.5\height}{\subcaptionbox{weak correlation}{\includegraphics[height= \figTwoHeight in]{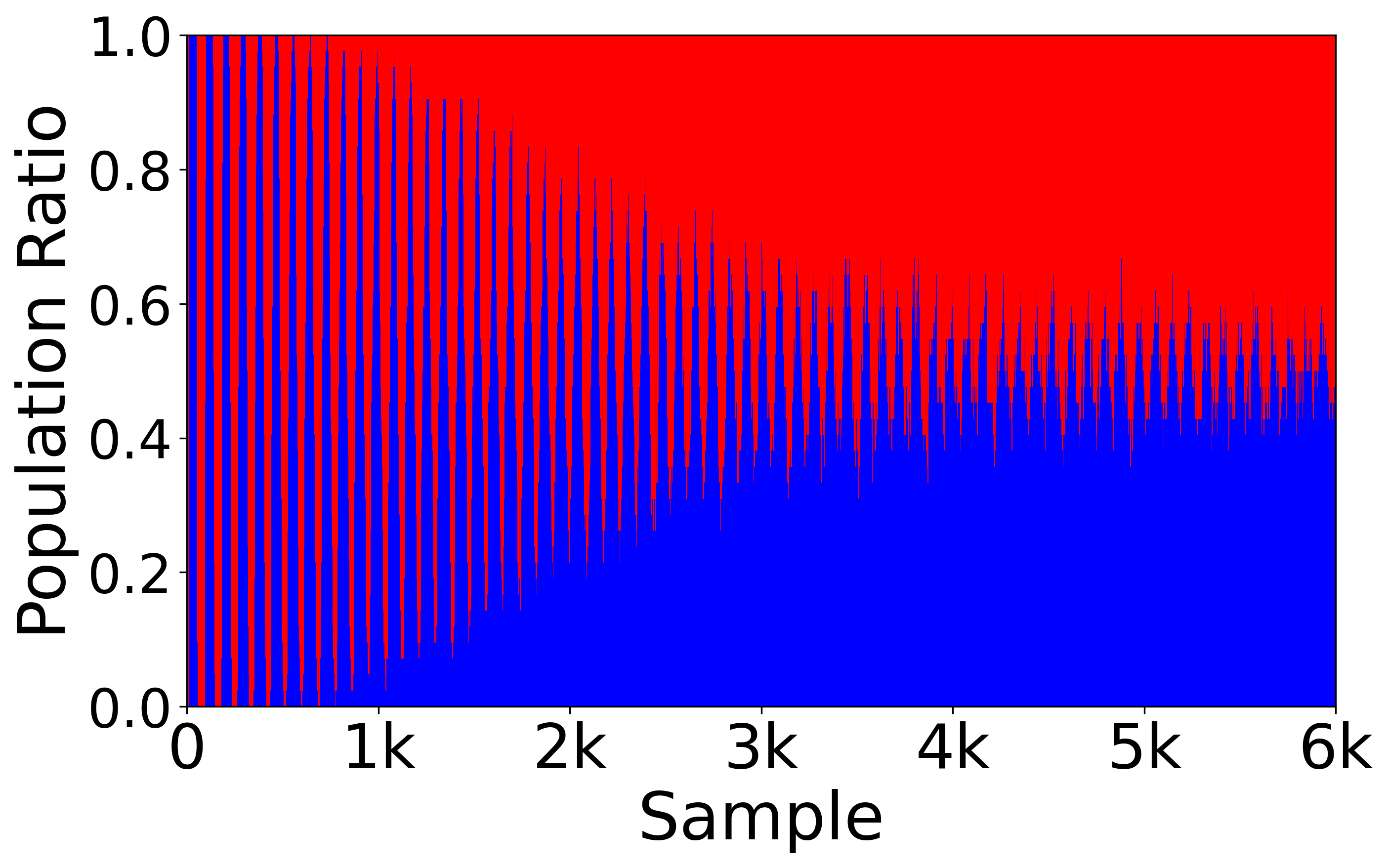}}}
    \hfill%
    \raisebox{-0.5\height}{\subcaptionbox{medium correlation}{\includegraphics[height=\figTwoHeight in]{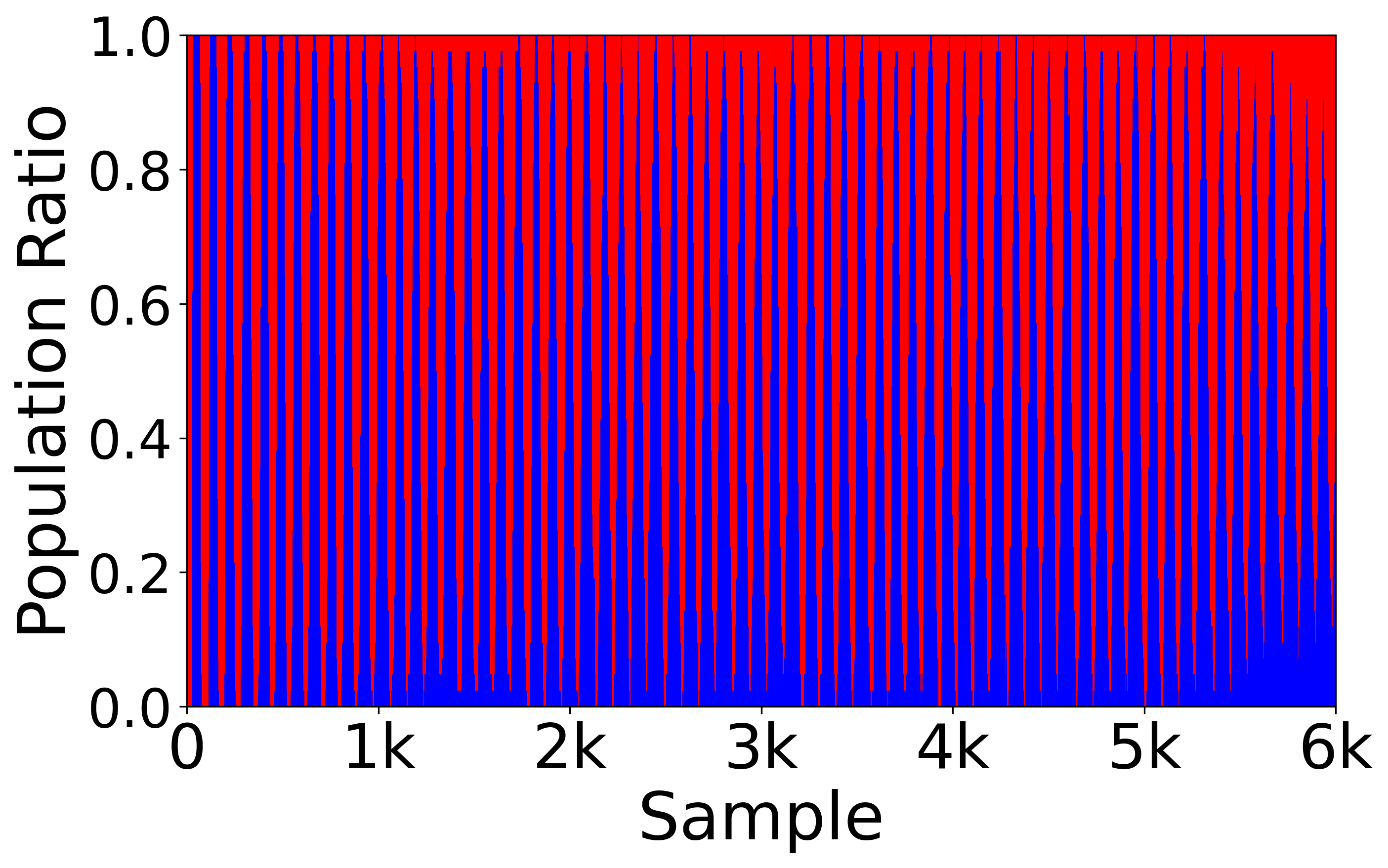}}}
    \hfill%
    \raisebox{-0.5\height}{\subcaptionbox{strong correlation}{\includegraphics[height=\figTwoHeight in]{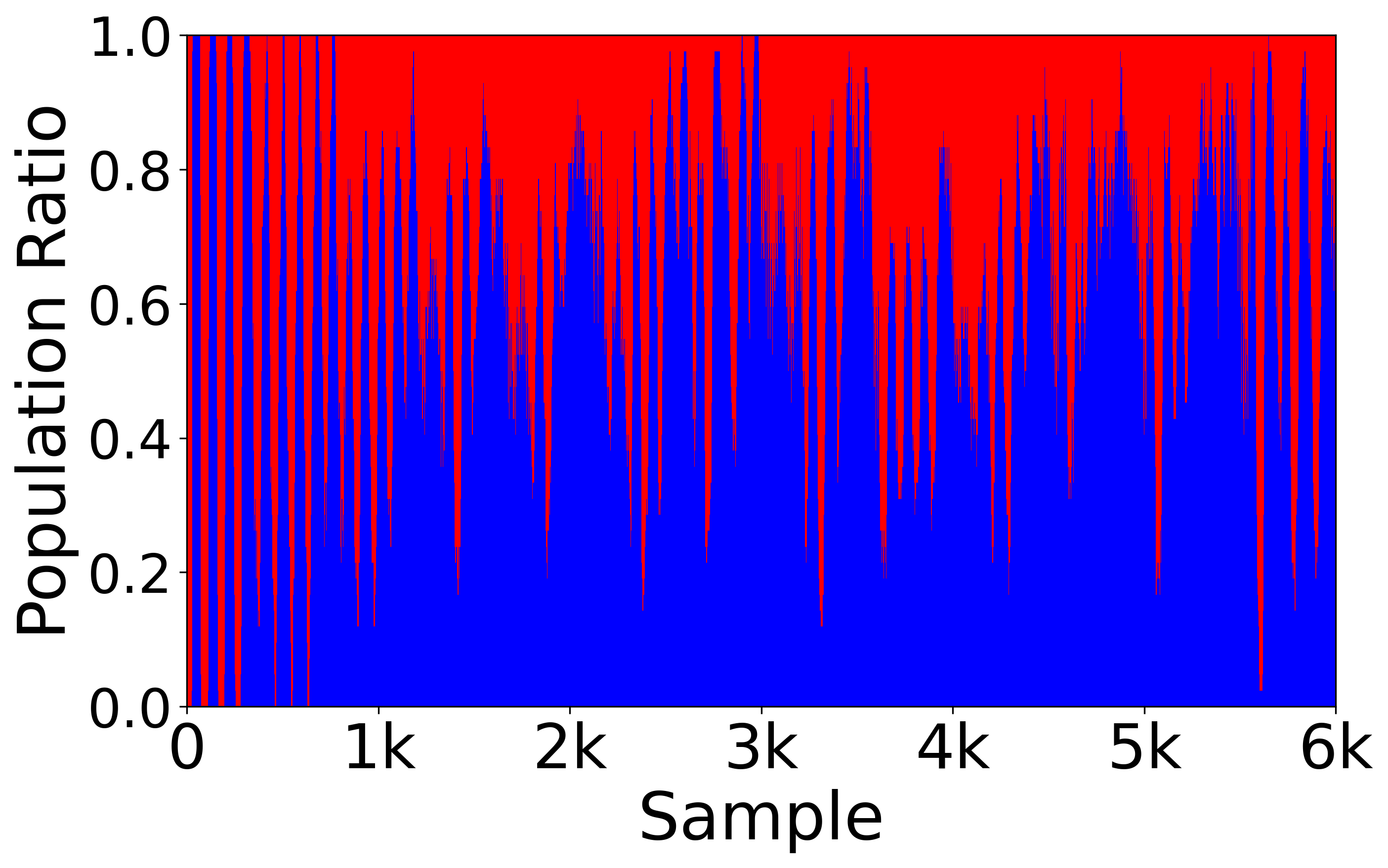}}}
\caption{a) 49 Kilobots on a grid are detected in either blue or red. b) Sorted distribution of switching numbers of 4 different repetitions. The right-most histogram shows the ensemble distribution. c) The composition of the population in red and blue versus sample time, for a long experiment showing the collective gets asynchronous over time. d-f) Kilobots synchronizing with low, medium, and high correlation strength, resulting in asynchronous, synchronous, and super-stable behaviors.}
\label{fig:phase_single}
\end{figure*}
The last aspect of individuality we discuss in this paper is the natural frequency of oscillating robots, due to the small differences in the internal clocking frequency. Natural frequency in its general meaning relates to a big family of periodic behavior, the natural frequency of a pendulum, any structure such as bridges, oscillators, or simply clocks. The microprocessor inside a robot, specifically the Kilobot, relies on its internal (or external) clock to perform the operations. These clocks have a nominal frequency with a minuscule tolerance. We investigate if, and to what extent Kilobots are heterogeneous in their frequency. Also, following an experiment, we show that the heterogeneity in the internal frequency contributes to asynchrony in collectives.
%
%
%
To measure the heterogeneity in the natural frequency of Kilobots, we program 49 robots to change their LED color (alternating between red and blue) every 30 interrupt pulses of the internal clock. The algorithm is as follows. Each robot counts the number of pulses it receives from the internal clock, resembling the phase ($\phi$) of the robot. We also define two states for robots associated with red and blue. If the phase is below 30, the robot shows its red LED, otherwise blue. The counter (and the phase) resets if it surpasses 60. To an observer of the robot, the LED alternates between red and blue almost every second. We locate all the robots on a lattice (Fig.~\ref{fig:phase_single}-a) and initiate their program exactly at the same time, by sending a broadcast message (minor individual deviations cannot be excluded but the results indicate good initial synchrony). During the experiment, no message is interchanged between robots. We recorded and post-processed the video of robots and labeled each robot by detecting the color of its LED.
\par
We count the number of times each robot switches and illustrated the distribution in Fig.~\ref{fig:phase_single}-b, where we observe the persistent heterogeneity in natural frequencies. In Fig.~\ref{fig:phase_single}-c, we show the population ratio of robots in either red or blue states, summing to one. At each time step, we count the ratio of robots with blue LED and drew a vertical line in blue with a height equal to the ratio. The rest of the population is shown by the upper red line. A~full blue or red line denotes the full synchronous collective state, which is the case for the initial state of the collective. Over time, the collective deviates from the synchrony and reaches a point where the population is divided into equal halves, for example at the 2000-th time step. Based on this result, we argue that asynchrony happens naturally in these systems, resulting in delays, and more complex dynamics on the network. 
There are two common approaches addressing the asynchrony in distributed artificial systems: to leave the system as is, and study how the delay affects the dynamics on the network~\citep{seuret2008consensus, tsianos2012impact}; or to push the system toward a synchronous collective state via the interaction of agents, which is a complex behavior~\citep{kuramoto1984chemical, strogatz2004sync, ceron2023diverse}, or as done, for example, in sensor networks~\citep{sundararaman2005clock,degesys2007desync}. We examined the synchronization in Kilobots, following a Kuramoto model~\citep{kuramoto1984chemical}, and show the collective behavior for different coupling strengths in~Fig.~\ref{fig:phase_single}-d-f.
Neither a weak (Fig.~\ref{fig:phase_single}-d) nor a too-strong correlation strength (Fig.~\ref{fig:phase_single}-f) result in a synchronous state. The optimal value to achieve synchrony lies somewhere in the middle of the previous two (Fig.~\ref{fig:phase_single}-e).
%
%
%
\section{Conclusion}
Inspired by studies on natural systems and the complex behavior caused by inter-individual variations, we attempt to shed some light on the concept of individuality in swarm robotics. We found that individuality in robot systems in general, and swarm robots in particular, are often overlooked. We argue that robots have agent-specific persistent features, that are characteristic parts of them. These natural differences are either assumed to be ``noise" or error and hence provoke solutions like calibration (as an offline solution) or regulation using feedback control (as an online solution). We argue that there is some useful information in the variations which can be exploited to make more accurate models and hence predictions. We also showed that robots develop individuality over the course of experiments, and thus calibration is not always a lasting solution. Also, regulating the errors comes with the price of a feedback signal, which is usually too costly for minimal swarm robots.
 \par
Furthermore, dropping individuality as a dimension of the problem space will lead to increased uncertainty in the model.
We observe, report, and measure the heterogeneity in motion, sensing, and frequency of real Kilobots, and show that the robots have agent-specific, persistent, non-zero mean biases.
With the accurate model for heterogeneity in heading bias, we scaled up our studies and showed how different robots vary in their performance. Our results provide evidence that calling inter-individual variations a bug or a flaw is not always true. Our results show a counter-intuitive comparison of the perfect and biased robots, with the perfect robot being outperformed by biased ones in some tasks.
Besides, the new perspective opens space for new insights to be gained from these complex systems.
 %
 %
For future works, we aim to extend our case study beyond Kilobots, and to investigate the effect of heterogeneity in collective tasks, for instance, how collectives with different levels of diversity compare in their performance. These tasks will not be limited to scenarios involving motion (e.g., collective motion), but also collective decision-making, perception, and synchronization are other examples of collective behavior whose complexity might stem from inter-individual variations.
\section{Acknowledgment}
This work is funded by the Deutsche Forschungsgemeinschaft (DFG, German Research Foundation) under Germany’s Excellence Strategy – EXC 2002/1 ``Science of Intelligence” – project number 390523135.
\footnotesize
\bibliographystyle{apalike}
\bibliography{example} 
\end{document}